\documentclass[Afour,sageh,times]{sagej}

\usepackage{moreverb,url}
\usepackage{subcaption}
\usepackage{gensymb}
\usepackage{multirow}
\usepackage{graphicx}
\usepackage{pifont}

\usepackage{booktabs}

\usepackage[colorlinks,bookmarksopen,bookmarksnumbered,citecolor=red,urlcolor=red]{hyperref}

\newcommand\BibTeX{{\rmfamily B\kern-.05em \textsc{i\kern-.025em b}\kern-.08em
T\kern-.1667em\lower.7ex\hbox{E}\kern-.125emX}}

\def\volumeyear{2024}

\begin{document}

\runninghead{Carmichael, et al.}

\title{Dataset and Benchmark: Novel Sensors for Autonomous Vehicle Perception}

\author{Spencer Carmichael\affilnum{1}, Austin Buchan\affilnum{1}, Mani Ramanagopal\affilnum{1*}, Radhika Ravi\affilnum{1*}, Ram Vasudevan\affilnum{1,2} and Katherine A. Skinner\affilnum{1}}

\affiliation{\affilnum{1}Robotics Department, University of Michigan, Ann Arbor, MI 48109 USA\\
\affilnum{2}Department of Mechanical Engineering, University of Michigan, Ann Arbor, MI 48109 USA\\
\affilnum{*}M. Ramanagopal and R. Ravi contributed to this work while employed at University of Michigan.
}

\corrauth{Spencer Carmichael, 2505 Hayward St., Ann Arbor, MI 48109}
\email{specarmi@umich.edu}

\begin{abstract}

Conventional cameras employed in autonomous vehicle (AV) systems support many perception tasks, but are challenged by low-light or high dynamic range scenes, adverse weather, and fast motion. Novel sensors, such as event and thermal cameras, offer capabilities with the potential to address these scenarios, but they remain to be fully exploited. This paper introduces the Novel Sensors for Autonomous Vehicle Perception (NSAVP) dataset to facilitate future research on this topic. The dataset was captured with a platform including stereo event, thermal, monochrome, and RGB cameras as well as a high precision navigation system providing ground truth poses. The data was collected by repeatedly driving two $\sim$8 km routes and includes varied lighting conditions and opposing viewpoint perspectives. We provide benchmarking experiments on the task of place recognition to demonstrate challenges and opportunities for novel sensors to enhance critical AV perception tasks. To our knowledge, the NSAVP dataset is the first to include stereo thermal cameras together with stereo event and monochrome cameras. The dataset and supporting software suite is available at: \url{https://umautobots.github.io/nsavp}.

\end{abstract}

\keywords{Data Sets for SLAM, Localization, Mapping, Sensor Fusion, Computer Vision for Transportation}

\maketitle

\section{Introduction}

Frame-based visible spectrum cameras are widely utilized with autonomous vehicles (AVs) due to their relatively low cost, high spatial resolution, and proven suitability for a variety of perception tasks. However, these sensors are challenged by low-light and adverse weather conditions and suffer from several issues including motion blur, limited frame rates, and low dynamic range \cite{zhang_perception_2023}.

Novel sensors, including event and thermal cameras, present advantages that may aid in overcoming these limitations. Event cameras feature high temporal resolution, negligible motion blur, high dynamic range, low power requirements, and low latency \cite{gallego_event-based_2022}. Thermal cameras sensitive to the Long Wave Infrared (LWIR) spectrum can operate in the absence of visible light and are robust to changes in illumination and visual obscurants such as dust, smoke and fog \cite{judd_automotive_2019}. However, these novel sensors also present their own challenges. Event cameras represent a new paradigm that is not directly compatible with existing computer vision algorithms. Affordable (uncooled microbolometer) thermal cameras suffer from rolling shutter distortions, significant motion blur, and fixed pattern noise, the latter of which necessitates non-uniformity corrections (NUCs) that may require image interruptions \cite{torres_kalman_2003}. 

\begin{figure*}
\centering
\includegraphics[width=0.75\textwidth]{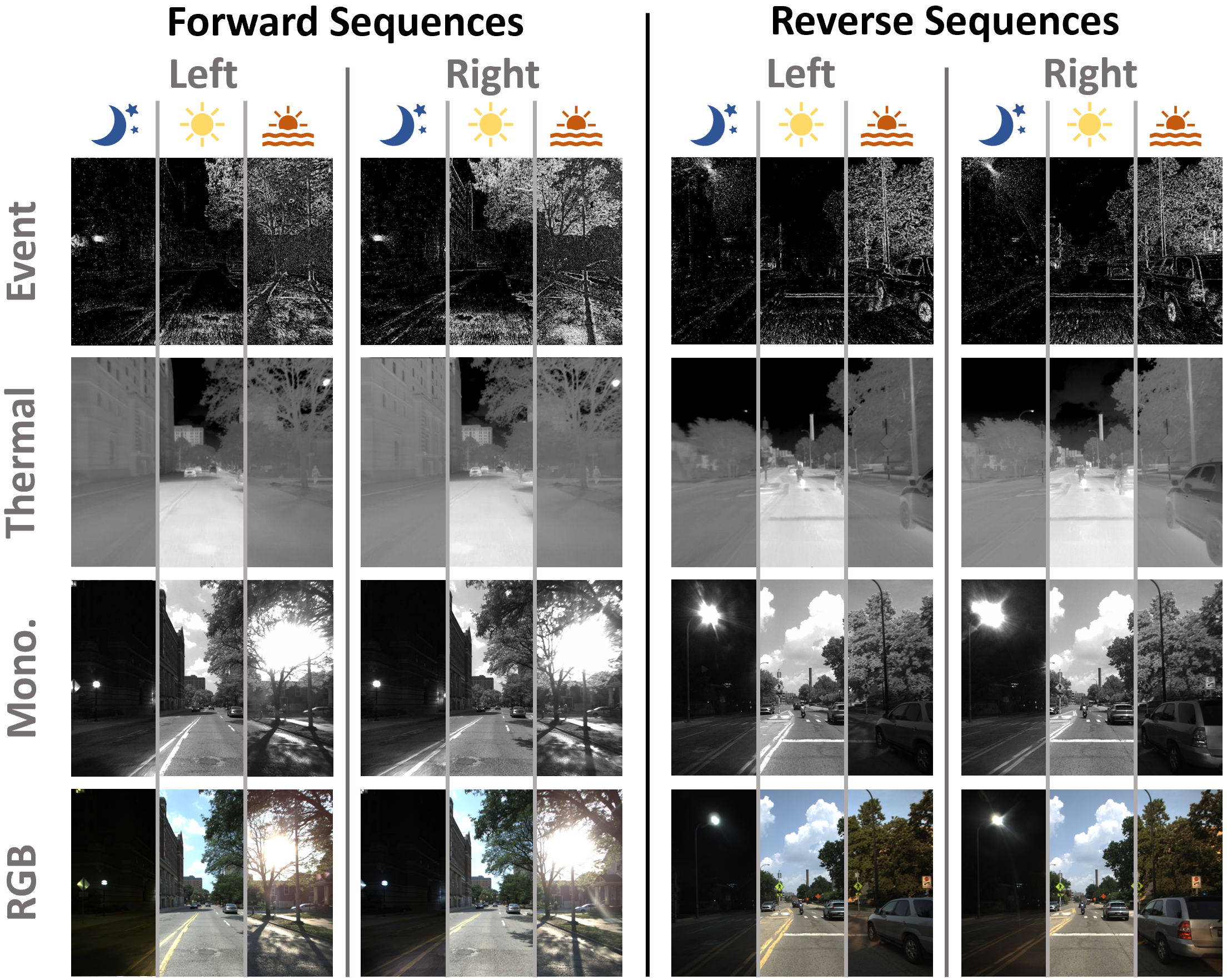}
\caption{Example data from each camera across six sequences capturing two opposing driving directions under three lighting conditions: night, afternoon, and sunset. For each camera, a composite image formed from the three lighting conditions is shown, with event data represented as a time surface \cite{gallego_event-based_2022}. All the data is from approximately the same location as determined by the provided ground truth poses. These examples highlight the robustness of thermal cameras to illumination changes and of thermal and event cameras to direct sunlight, which produces saturation and blooming in conventional cameras.}  
\label{figure:examples}
\end{figure*}

In this paper, we present the Novel Sensors for Autonomous Vehicle Perception (NSAVP) dataset, which features conventional RGB and monochrome stereo cameras alongside stereo event and stereo thermal cameras. Sequences were collected along two $\sim$8 km routes. Both routes were driven in the forward and reverse directions to facilitate opposing viewpoint place recognition evaluation. Additionally, one route was driven at three times of day (afternoon, sunset, and night) to capture both benign and challenging lighting conditions. We present benchmarking experiments to evaluate place recognition under illumination changes. The spanned combinations of modality, viewpoint, and lighting are visualized in Fig.~\ref{figure:examples}. We hope the dataset will enable further research that compares or combines conventional and novel sensors for AV perception.

\section{Related Work}\label{section:related_work}

In recent years, there has been increasing interest and availability in AV datasets with non-traditional sensors, including thermal cameras and event cameras. The majority of AV datasets including thermal cameras address pedestrian and object detection \cite{miron_evaluation_2015, hwang_multispectral_2015, takumi_multispectral_2017, noauthor_free_nodate, XU2019199, tumas_pedestrian_2020, valverde_there_2021, farooq_evaluation_2023} or semantic segmentation \cite{ha_mfnet_2017, vertens_heatnet_2020, franchi_infraparis_2023} without supporting localization and mapping tasks. Early exceptions are the KAIST Day/Night \cite{choi_kaist_2018} and Brno Urban \cite{ligocki_brno_2020} datasets, which  include data from lidar and RTK GNSS systems serving as depth and pose ground truth, respectively. However, the KAIST Day/Night and Brno Urban datasets only provide monocular thermal cameras and do not include event or monochrome cameras.

Similarly, many earlier AV datasets including event cameras target tasks such as object detection \cite{sironi_hats_2018, de_tournemire_large_2020, perot_learning_2020}, image reconstruction \cite{scheerlinck_ced_2019, rebecq_high_2021}, and lane extraction \cite{cheng_det_2019}. The DDD17 dataset \cite{binas_ddd17_2017} was created primarily to investigate event-based end-to-end driving with vehicle control data, but it also includes GPS position data and overlapping sequences, making it the first dataset suitable for event-based place recognition evaluation. The DDD20 dataset \cite{hu_ddd20_2020} later expanded upon the DDD17 dataset, and concurrently the Brisbane-Event-VPR dataset \cite{fischer_event-based_2020} was released to allow large-scale event-based place recognition to be tested under varied illumination and weather conditions. The MVSEC dataset \cite{zhu_multi_2018}, which features multiple platforms, including a car, is the first dataset to include synchronized stereo event cameras. However, the event cameras have a low 0.09 MP resolution and narrow 10 cm baseline owing to the sensor rig being designed to fit small platforms. As a result, event-based stereo depth estimation is limited in the MVSEC AV sequences \cite{hadviger_feature-based_2021}. In response, the DSEC dataset \cite{gehrig_dsec_2021} is explicitly designed for event-based stereo depth estimation with higher resolution (0.31 MP) event cameras and a larger 51 cm baseline. However, while useful for that specific task, the DSEC dataset does not provide ground truth trajectories and is therefore unsuitable for visual odometry  or place recognition evaluation. The proposed NSAVP dataset also includes stereo event cameras with a wide baseline and additionally offers overlapping sequences and precise ground truth pose enabling both place recognition and visual odometry evaluation.

\begin{table*}[ht!]
\centering
\caption{Comparison with AV datasets which include event or thermal cameras and support localization and mapping tasks. For multi-platform datasets, the table reflects the AV sequences only. The proposed NSAVP dataset is the only to include each camera modality and to include stereo thermal cameras together with stereo event and monochrome cameras.}
\resizebox{0.95\textwidth}{!}{%
\begin{tabular}{@{}cccccc@{}}
\toprule \toprule
\multirow{2}{*}{\textbf{Dataset}}                  & \multicolumn{4}{c}{\textbf{Cameras (Megapixels $\times$ Quantity)}}                                                                                                                                                                                                &  \multirow{2}{*}{\textbf{\begin{tabular}[c]{@{}c@{}}Ground Truth\\ Pose/Position\end{tabular}}} \\ \cmidrule(lr){2-5}
                                                   & \textbf{Thermal} & \textbf{Event}                                                            & \textbf{Monochrome}                                                              & \textbf{RGB}                                                                                                                                                           &                                                                                                \\ \midrule
DDD17 \cite{binas_ddd17_2017}                      &                  & 0.09 $\times$ 1                                                           & 0.09 $\times$ 1$^\ast$                                                           &                                                                                                                                                                          & $\bigstar$                                                                                     \\ \midrule
MVSEC \cite{zhu_multi_2018}                        &                  & 0.09 $\times$ 2                                                           & \begin{tabular}[c]{@{}c@{}}0.09 $\times$ 2$^\ast$\\ 0.36 $\times$ 2\end{tabular} &                                                                                                                                                                   & $\bigstar$                                                                                     \\ \midrule
KAIST Day/Night \cite{choi_kaist_2018}             & 0.31 $\times$ 1  &                                                                           &                                                                                  & 1.23 $\times$ 2                                                                                                                                                   & $\bigstar$ $\bigstar$ $\bigstar$                                                               \\ \midrule
Brno Urban \cite{ligocki_brno_2020}                & 0.33 $\times$ 1  &                                                                           &                                                                                  & 2.30 $\times$ 4                                                                                                                                                   & $\bigstar$ $\bigstar$                                                                          \\ \midrule
DDD20 \cite{hu_ddd20_2020}                         &                  & 0.09 $\times$ 1                                                           & 0.09 $\times$ 1$^\ast$                                                           &                                                                                                                                                                   & $\bigstar$                                                                                     \\ \midrule
Brisbane-Event-VPR \cite{fischer_event-based_2020} &                  & 0.09 $\times$ 1                                                           &                                                                                  & \begin{tabular}[c]{@{}c@{}}0.09 $\times$ 1$^\ast$\\ 2.07 $\times$ 1\end{tabular}                                                                                                & $\bigstar$                                                                                     \\ \midrule
DSEC \cite{gehrig_dsec_2021}                       &                  & 0.31 $\times$ 2                                                           &                                                                                  & 1.56 $\times$ 2                                                                                                                                                   &                                                                                                \\ \midrule
ViViD++ \cite{lee_vivid_2022}                      & 0.33 $\times$ 1  & 0.31 $\times$ 1                                                           &                                                                                  & 1.31 $\times$ 1                                                                                                                                                   & $\bigstar$ $\bigstar$                                                                          \\ \midrule
STheReO \cite{yun_sthereo_2022}                    & 0.33 $\times$ 2  &                                                                           &                                                                                  & 0.72 $\times$ 2                                                                                                                                                   & $\bigstar$ $\bigstar$ $\bigstar$                                                               \\ \midrule
M3ED \cite{chaney_m3ed_2023}                       &                  & 0.92 $\times$ 2                                                           & 1.02 $\times$ 2                                                                  & 1.02 $\times$ 1                                                                                                                                                   & $\bigstar$ $\bigstar$                                                                          \\ \midrule
MS$^2$ \cite{shin_deep_2023}                          & 0.33 $\times$ 2  &                                                                           &                                                                                  & 5.01 $\times$ 2                                                                                                                                                   & $\bigstar$                                                                                     \\ \midrule
Stereo Visual Localization \cite{hadviger_stereo_2023}     &                  & 0.31 $\times$ 2                                                           &                                                                                  & 3.15 $\times$ 2                                                                                                                                                   & $\bigstar$ $\bigstar$ $\bigstar$                                                               \\ \midrule
MA-VIED \cite{mollica_ma-vied_2023}                &                  & 0.31 $\times$ 1                                                           & 2.30 $\times$ 1                                                                  &                                                                                                                                                                   & $\bigstar$ $\bigstar$ $\bigstar$                                                               \\ \midrule
NTU4DRadLM \cite{zhang_ntu4dradlm_2023}            & 0.33 $\times$ 1  &                                                                           &                                                                                  & 0.31 $\times$ 1                                                                                                                                                   & $\bigstar$ $\bigstar$                                                                          \\ \midrule
ECMD \cite{chen_ecmd_2023}                         & 0.33 $\times$ 1  & \begin{tabular}[c]{@{}c@{}}0.09 $\times$ 2\\ 0.31 $\times$ 2\end{tabular} &                                                                                  & \begin{tabular}[c]{@{}c@{}}0.09 $\times$ 2$^\ast$\\ 2.30 $\times$ 2\end{tabular}                                                                                  & $\bigstar$ $\bigstar$ $\bigstar$                                                               \\ \midrule
\textbf{NSAVP}                                              & 0.33 $\times$ 2  & 0.31 $\times$ 2                                                           & 1.56 $\times$ 2                                                                  & 1.25 $\times$ 2                                                                                                                                                   & $\bigstar$ $\bigstar$ $\bigstar$                                                               \\ \bottomrule \bottomrule
\multicolumn{6}{l}{$\bigstar$: Non-differential GNSS or SLAM-derived, $\bigstar$$\bigstar$: RTK or PPK GNSS Position, $\bigstar$$\bigstar$$\bigstar$: RTK or PPK GNSS/IMU 6-DoF Pose} \\
\multicolumn{6}{l}{$^\ast$: Frame captured by DAVIS event camera}
\end{tabular}%
}
\label{table:dataset_comparison}
\end{table*}

In the past two years, several AV datasets have been published with thermal or event cameras and ground truth trajectories. ViViD++ \cite{lee_vivid_2022} is the first to have both modalities, with a single camera of each, followed by ECMD \cite{chen_ecmd_2023}, which includes a monocular thermal camera together with stereo event cameras. STheReO \cite{yun_sthereo_2022}, and subsequently MS$^2$ \cite{shin_deep_2023}, are the first to provide asynchronous and synchronous stereo thermal camera data, respectively. NTU4DRadLM \cite{zhang_ntu4dradlm_2023} is the first to pair a thermal camera with a 4D radar system. Similar to MVSEC, the M3ED \cite{chaney_m3ed_2023} dataset includes a stereo event camera attached to a single sensor rig placed on multiple platforms of varied sizes. Again, the small baseline (12 cm) limits stereo depth estimation in typical AV environments, although this is mitigated by the cameras' high 0.92 MP resolution \cite{chaney_m3ed_2023}. Similar to DDD17 and DDD20, MA-VIED \cite{mollica_ma-vied_2023} provides vehicle control data alongside a monocular event camera, but includes a higher resolution event camera, higher quality pose ground truth, and the addition of a monochrome camera, IMU, and wheel odometer. Finally, the Stereo Visual Localization dataset \cite{hadviger_stereo_2023} addresses the gap left by DSEC by capturing AV sequences with a wide baseline stereo event camera and ground truth poses from a dual antenna RTK GNSS/IMU system. While these datasets encompass a variety of sensor suites, none include stereo thermal cameras together with stereo event or monochrome cameras.

Table~\ref{table:dataset_comparison} gives a comparison with all AV datasets that include event or thermal cameras and support localization and mapping tasks. Our contributed dataset is the first to include stereo thermal cameras together with stereo event and monochrome cameras, and it provides precise time synchronization between all sensors and highly accurate ground truth poses. Thermal and event cameras complement each other in terms of robustness to motion and low-light conditions making the fusion of these modalities promising. Additionally, monochrome cameras are more sensitive than RGB cameras and do not suffer from debayering artifacts making them a better comparison point to novel sensors in low-light conditions. Moreover, by repeatedly driving large-scale routes in opposing directions, our dataset is uniquely able to support opposing viewpoint place recognition evaluation. Finally, we note that ground truth object bounding boxes and pixelwise semantic labels could be added to our dataset in the future, similar to the extensions of DDD17 \cite{alonso_ev-segnet_2019}, MVSEC \cite{hu_v2e_2021}, and DSEC \cite{avidan_ess_2022}.

\section{Platform and Sensors}

\begin{figure}
\includegraphics[width=1.0\columnwidth]{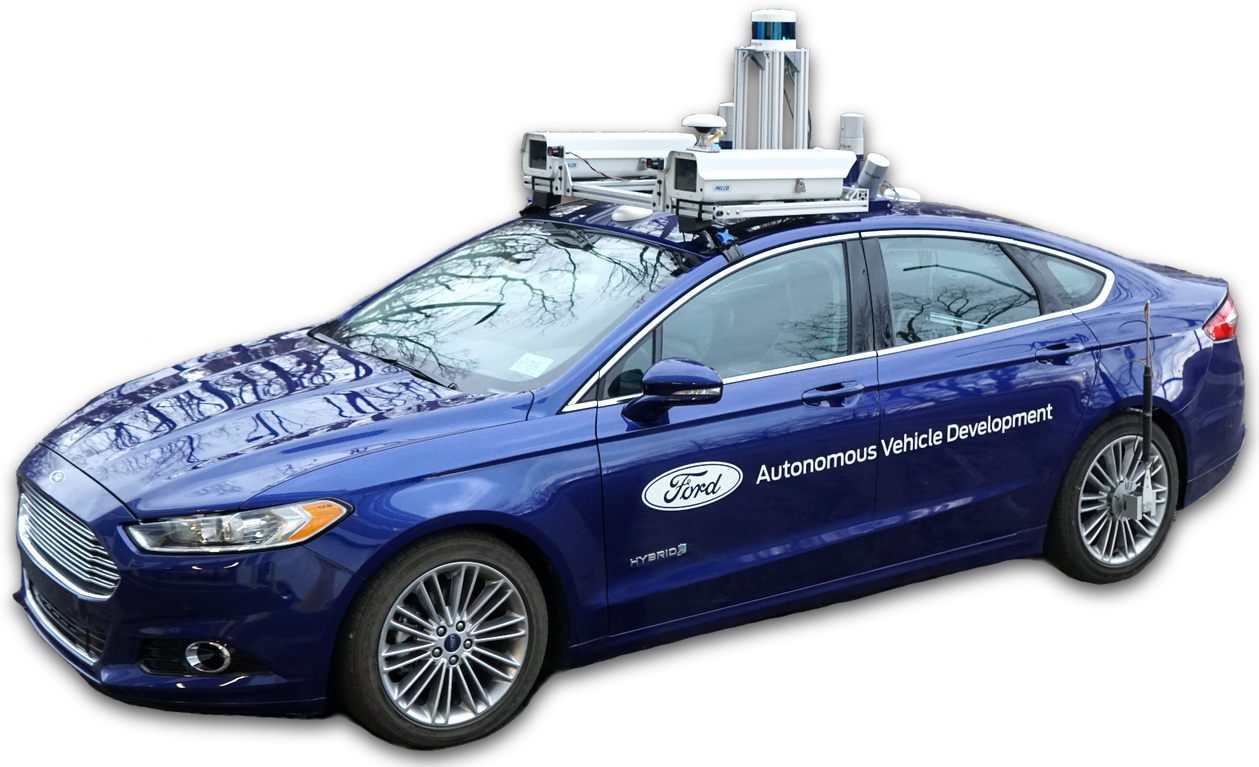}
\includegraphics[width=1.0\columnwidth]{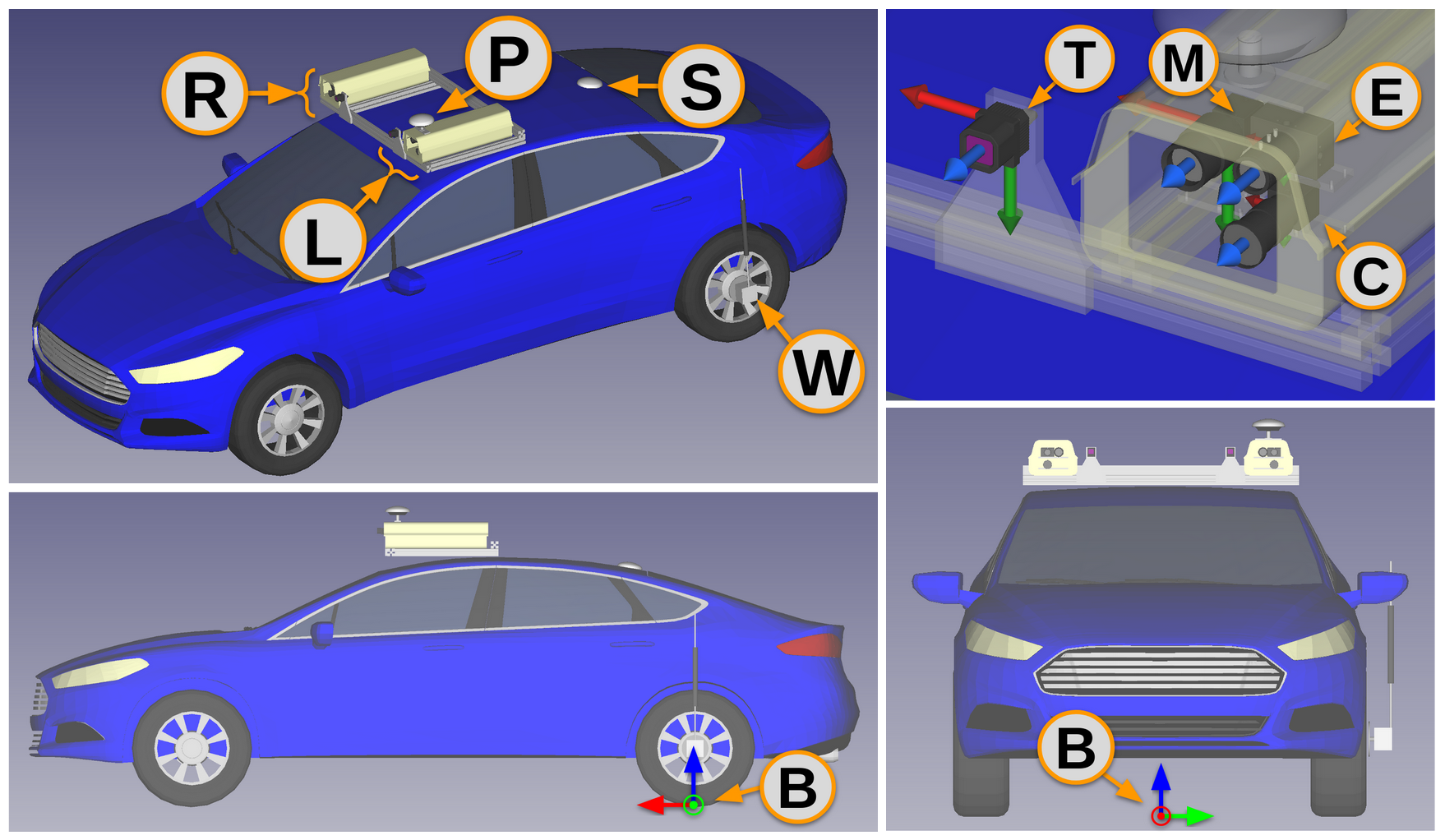}
\caption[Vehicle Sensor Platform. The labeled vehicle diagram shows the position and orientation of the symmetric \textbf{L}eft and \textbf{R}ight groups of vision sensors, \textbf{P}rimary and \textbf{S}econdary POS-LV 420 antennas, and POS-LV 420 \textbf{W}heel encoder. The sensor detail diagram shows the left sensor group, with a \textbf{T}hermal camera, \textbf{E}vent camera, \textbf{M}onochrome camera, and \textbf{C}olor (RGB) camera. The left and front vehicle views show the location of the system reference \textbf{B}ase link frame. Coordinate frames use red, green, and blue arrows for x, y, and z axes respectively.]{Vehicle Sensor Platform\endnote{CAD sources:\newline Monochrome and RGB camera models \newline\url{https://www.flir.com/support-center/iis/machine-vision/knowledge-base/technical-documentation-bfs-gige/}\newline Thermal camera model\newline\url{https://www.flir.com/support/products/adk/?pn=ADK&vn=40640U050-6PAAX\#Downloads}\newline Event camera model\newline\url{https://grabcad.com/library/davis346-event-camera-1}\newline Ford Fusion model (released with the Ford Multi-AV Seasonal Dataset \cite{ford_multi_av})\newline\url{https://github.com/Ford/AVData/blob/master/fusion_description/meshes/Ford_fusion.obj}}. The labeled vehicle diagram shows the position and orientation of the symmetric \textbf{L}eft and \textbf{R}ight groups of vision sensors, \textbf{P}rimary and \textbf{S}econdary POS-LV 420 antennas, and POS-LV 420 \textbf{W}heel encoder. The sensor detail diagram shows the left sensor group, with a \textbf{T}hermal camera, \textbf{E}vent camera, \textbf{M}onochrome camera, and \textbf{C}olor (RGB) camera. The left and front vehicle views show the location of the system reference \textbf{B}ase link frame. Coordinate frames use red, green, and blue arrows for x, y, and z axes respectively.}
\label{figure:platform_labeled}
\end{figure}

The proposed dataset was collected using a Ford Fusion platform with a custom sensor suite as pictured in Fig.~\ref{figure:platform_labeled}. The sensor suite includes stereo pairs of each camera detailed in Table~\ref{table:camera_specs}, a Protempis Thunderbolt GM200 grandmaster clock, and an Applanix POS-LV 420 navigation system featuring a dual antenna GNSS receiver, IMU, and wheel encoder. We use the GM200 for time synchronization, as described in section~\ref{section:camera_time_sync}, and the POS-LV 420 to obtain 200 Hz ground truth poses of the vehicle's `base link' frame, as described in section~\ref{section:ground_truth}. An on-board compute and network system performs sensor initialization, configuration, and data logging. The compute system with a Ryzen 9 5950X 3.4GHz 16-Core CPU, 64GB DDR4 RAM, 4TB NVMe solid-state drive, and 10Gbit Ethernet ports was sufficient to handle saving the 3Gbps+ raw data stream to disk.

\begin{table*}[]
\centering
\caption{Specifications of the stereo cameras included in the sensor suite.}
\resizebox{1.00\textwidth}{!}{
\begin{tabular}{@{}cccccccc@{}}
\toprule \toprule
\textbf{Type}                                                                   & \textbf{Camera Model}                                                          & \textbf{Lens Model}     & \textbf{\begin{tabular}[c]{@{}c@{}}Stereo\\ Baseline\\ (meters)\end{tabular}} & \textbf{\begin{tabular}[c]{@{}c@{}}FOV\\ (H $\times$ V)\end{tabular}} & \textbf{\begin{tabular}[c]{@{}c@{}}Resolution\\ (W $\times$ H)\end{tabular}} & \textbf{\begin{tabular}[c]{@{}c@{}}Rate\\ (Hz)\end{tabular}} & \textbf{\begin{tabular}[c]{@{}c@{}}Frame\\ Readout\end{tabular}} \\ \midrule
\begin{tabular}[c]{@{}c@{}}Monochrome\\ (CMOS)\end{tabular}                     & \begin{tabular}[c]{@{}c@{}}FLIR BFS-PGE-16S2M\\ (FLIR Blackfly S GigE)\end{tabular} & Computar A4Z2812CS-MPIR & 0.90                                                                          & 70\degree $\times$ 55\degree                                           & 1440 $\times$ 1080                                                           & 20.14                                                        & Global                                                           \\ \midrule
\begin{tabular}[c]{@{}c@{}}RGB\\ (CMOS)\end{tabular}                            & \begin{tabular}[c]{@{}c@{}}FLIR BFS-PGE-50S5C\\ (FLIR Blackfly S GigE)\end{tabular} & Fujinon HF6XA-5M        & 1.00                                                                          & 70\degree $\times$ 60\degree                                           & 1224 $\times$ 1024                                                           & 20.14                                                        & Global                                                           \\ \midrule
\begin{tabular}[c]{@{}c@{}}Thermal\\ (Uncooled VOx Microbolometer)\end{tabular} & \begin{tabular}[c]{@{}c@{}}FLIR 40640U050-6PAAX\\ (FLIR ADK)\end{tabular}      & NA (Integrated)         & 0.64                                                                          & 50\degree $\times$ 40\degree                                           & 640 $\times$ 512                                                             & 60.42                                                        & Rolling                                                          \\ \midrule
\begin{tabular}[c]{@{}c@{}}Event\\ (DVS)\end{tabular}                           & Inivation DVXplorer                                                            & NA (Included with Camera)           & 1.00                                                                          & 70\degree $\times$ 50\degree                                           & 640 $\times$ 480                                                             & NA                                                           & NA                                                               \\ \bottomrule \bottomrule
\multicolumn{8}{l}{NA: Not applicable}
\end{tabular}%
}
\label{table:camera_specs}
\end{table*}

% Vehicle platform image with labeled CAD diagrams

\begin{figure}
\centering
\includegraphics[width=1.0\columnwidth]{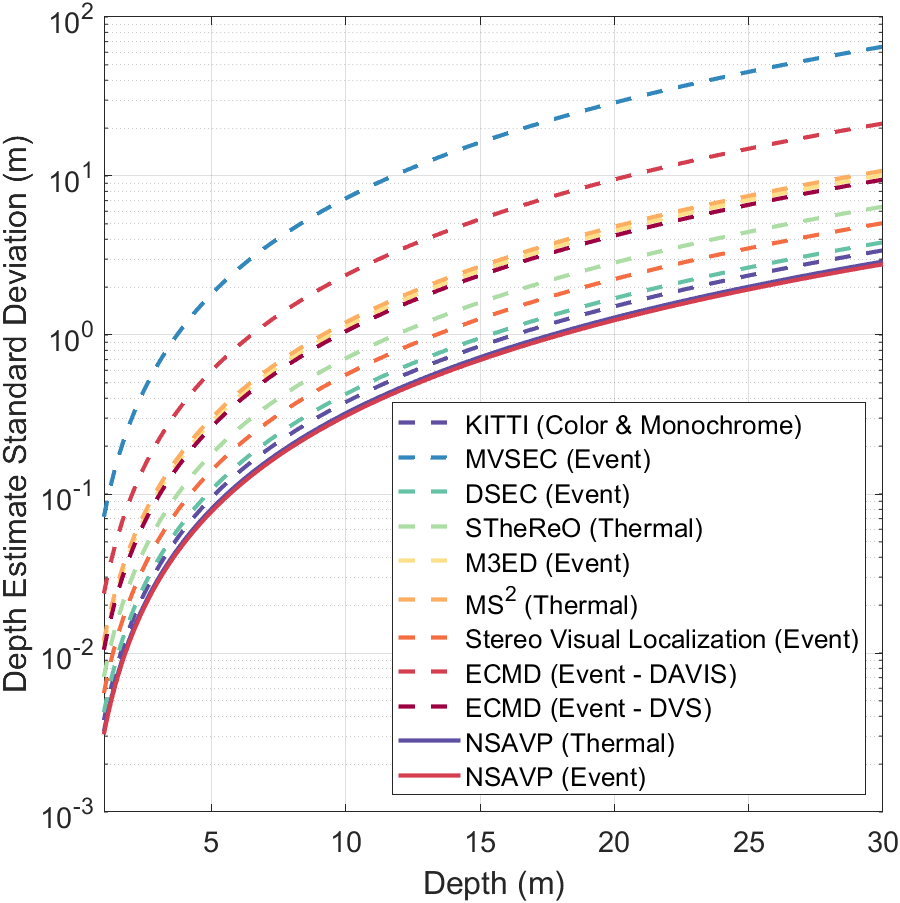}
\caption{Comparison of thermal and event stereo depth estimate uncertainty across AV datasets assuming the error model described in \cite{matthies_error_1987}. Conventional stereo cameras from the established KITTI dataset \cite{geiger_vision_2013} are also included for reference. NSAVP achieves the lowest predicted depth uncertainty.}  
\label{figure:stereo_error}
\end{figure}

Figure~\ref{figure:platform_labeled} shows the locations of the sensors on the vehicle and the relevant coordinate frames. The visible spectrum cameras are housed in enclosures on either side of the car in a mirrored configuration and are clustered tightly together within each enclosure to minimize parallax. The thermal cameras sit outside of the enclosures, as the enclosure glass is opaque in the LWIR spectrum. In selecting stereo baselines we considered triangulation error and field of view (FOV) overlap. We selected relatively large baselines to reduce triangulation error, placed the thermal cameras on the inside due to their narrower field of view, and placed the event cameras on the outside due to their lower resolution. Figure~\ref{figure:stereo_error} compares depth estimate uncertainty between the thermal and event stereo pairs included in our dataset and the datasets listed in Table~\ref{table:dataset_comparison}. Our configuration achieves the lowest predicted depth uncertainty.  

\section{Data Collection}

\subsection{Camera Settings}

The monochrome and RGB cameras are set to utilize auto-exposure and auto-gain such that exposure is increased first, up to a maximum of 15 ms, and then gain is increased, up to a maximum of 18 dB. We enable auto-white balancing and gamma correction in the RGB cameras and disable gamma correction in the monochrome cameras. To reduce bandwidth and storage requirements, we enable 2$\times$2 pixel binning in the RGB cameras, resulting in the resolution given in Table~\ref{table:camera_specs}. The event cameras' bias sensitivity is set to the default preset.

The thermal cameras are set to perform NUCs automatically\endnote{The NUC discussed in this section involves a shutter which closes to present a uniform thermal signal to the sensor array. Such a NUC is referred to as calibration-based. The thermal cameras also continuously perform a scene-based NUC.}. Each NUC produces a $\sim$0.5 second period of dropped frames. NUCs occur every $\sim$3-5 minutes throughout the dataset and are most frequent in the afternoon sequences. Notably, there are no instances in the dataset in which a NUC occurs simultaneously in the left and right thermal cameras. This uninterrupted availability of thermal images could allow, for example, a stereo thermal odometry method to temporarily propagate scale with one camera during a NUC on the other.

\subsection{Camera Time Synchronization}\label{section:camera_time_sync}

\begin{figure}
\centering
\includegraphics[width=1.0\columnwidth]{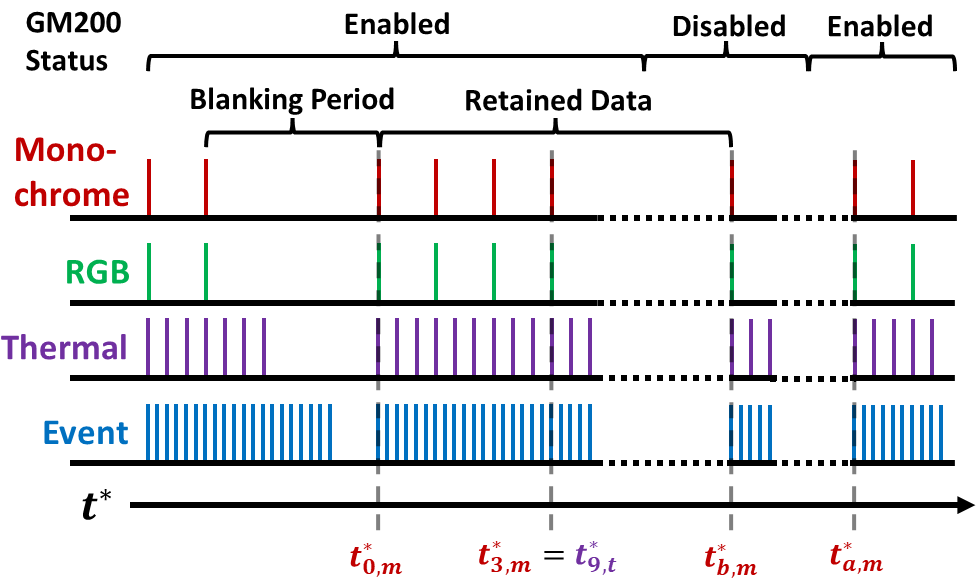}
\caption{Visualization of time synchronization signals sent to each camera modality. For simplicity, single vertical lines are used to represent the square wave edges (rising or falling) each sensor responds to. The durations of the blanking period and frequency of the event camera signal are not to scale.}  
\label{figure:sync}
\end{figure}

To achieve synchronized image capture and timestamps we trigger the cameras with an Arty Z7 FPGA development board connected to the cameras through a custom adapter PCB. The RGB and monochrome cameras receive a nominal 20 Hz square wave and the thermal cameras receive a nominal 60 Hz square wave such that every RGB and monochrome image, and every third thermal image, are captured simultaneously. Specifically, the signals trigger the start of exposure in the RGB and monochrome cameras and the start of readout in the thermal cameras. The event cameras receive a 10 kHz square wave, which advances their timestamps on each falling edge. These signals are visualized in Fig.~\ref{figure:sync}.

To synchronize the timestamps across all modalities we employ a careful strategy. The GM200 is able to synchronize the monochrome and RGB camera clocks with International Atomic Time (TAI) via the Precision Time Protocol (PTP) with nanosecond accuracy, but it can operate only while stationary. We ensure the vehicle is parked at the beginning and end of each recorded sequence to allow the use of the GM200, and we disable the GM200 while in motion. We poll and record the PTP status of the cameras during data collection. The thermal and event cameras do not support PTP, so we apply a blanking period to each trigger signal while the GM200 is enabled to provide an association across all cameras. The blanking period duration varies to satisfy different interface requirements, but the end of the blanking period is aligned across all cameras and can be programmatically identified after collection in the recorded data. The blanking period appears as missing frames in the RGB, monochrome, and thermal cameras, and resets the event camera clock to start from zero at the end of the blanking period.

With this strategy we are able to correct all timestamps in post-processing. We define the aligned edge at the end of the blanking period as trigger 0. For each frame-based camera, we assign trigger indices to images by counting forward from trigger 0 while accounting for dropped frames. Let $t_{0,m}$ and $t_{0,m}^*$ denote the camera clock and TAI timestamps of the left monochrome image at trigger 0, respectively. The blanking period is applied while the GM200 is enabled and therefore $t_{0,m} \approx t_{0,m}^*$. Similarly, let $t_{a,m} \approx t_{a,m}^*$ denote the timestamp of a left monochrome image, with trigger index $a$, captured after the GM200 is re-enabled at the end of the sequence. The TAI timestamp of an image shortly before the GM200 is re-enabled can be estimated as $\tilde{t}_{b, m}^* = t_{a, m}^* - T (a - b)$, where $T$ is the effective trigger signal period derived from the camera clock timestamps. To correct for clock drift during the period where the GM200 is disabled, the TAI timestamps for all left monochrome images between trigger 0 and $b$ are estimated as follows:
\begin{align}\label{equation:timestamp_correction}
    \tilde{t}_{i, m}^* = t_{0,m}^* + (t_{i, m} - t_{0,m}^*)c_m, \quad & c_m = \frac{\tilde{t}_{b, m}^* - t_{0,m}^*}{t_{b, m} - t_{0,m}^*}
\end{align}
All images before trigger 0 and after trigger $b$ are dropped. To validate the above procedure, we repeat it with each RGB and monochrome camera and compare the resulting TAI timestamps at each trigger index. Across the dataset, the average disagreement is on the order of 10 $\mu$s and the maximum is 0.22 ms.

After the left monochrome timestamps are corrected, they are directly applied to all simultaneously triggered images and used to interpolate all nonsimultaneous image timestamps. That is, $\tilde{t}_{3i, t}^*=\tilde{t}_{i, m}^*$ where $\tilde{t}_{3i, t}^*$ denotes the TAI timestamp assigned to the left thermal image at trigger index $3i$, etc.

Given that the event camera clocks are reset at the end of the blanking period and are in lockstep with the FPGA's clock, the TAI timestamp of the $i_{th}$ event $t^*_{i, e}$ can be estimated from its camera clock timestamp $t_{i, e}$ as:
\begin{align}\label{equation:timestamp_correction}
    \tilde{t}_{i, e}^* = t_{0,m}^* + t_{i, e}c_e, \quad & c_e = \frac{t_{a, m}^* - t_{0,m}^*}{a \tilde{T}}
\end{align}
where $\tilde{T}$ is the nominal period of the monochrome camera trigger signal, as configured on the FPGA.

\subsection{Camera Calibration}\label{section:camera_calibration}

\begin{figure}
    \centering
    \begin{subfigure}{0.475\columnwidth}
        \centering
        \includegraphics[width=\columnwidth]{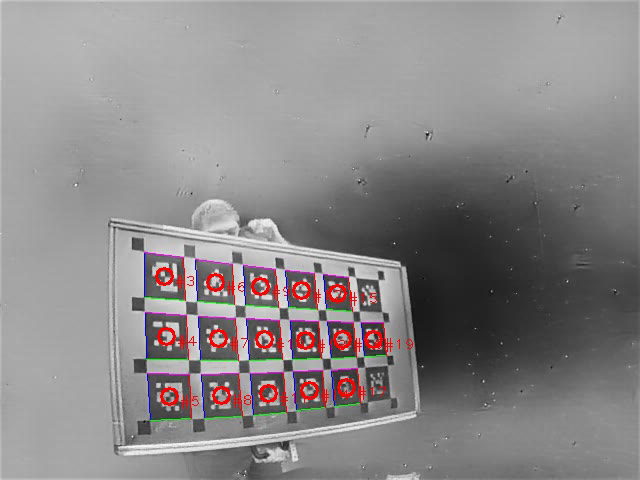}
        \caption{Event-based intensity reconstruction}
    \end{subfigure}
    \begin{subfigure}{0.475\columnwidth}  
        \centering 
        \includegraphics[width=\columnwidth]{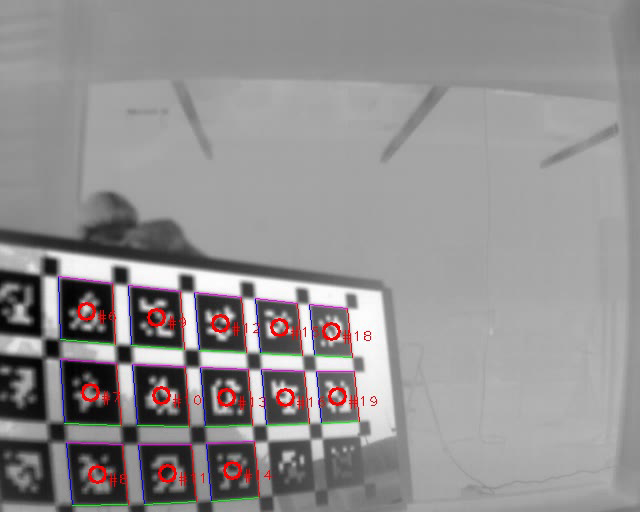}
        \caption{Thermal image truncated, inverted, and normalized to 8-bit}
        \label{subfigure:calibration_thermal}
    \end{subfigure}
    \begin{subfigure}{0.475\columnwidth}   
        \centering 
        \includegraphics[width=\columnwidth]{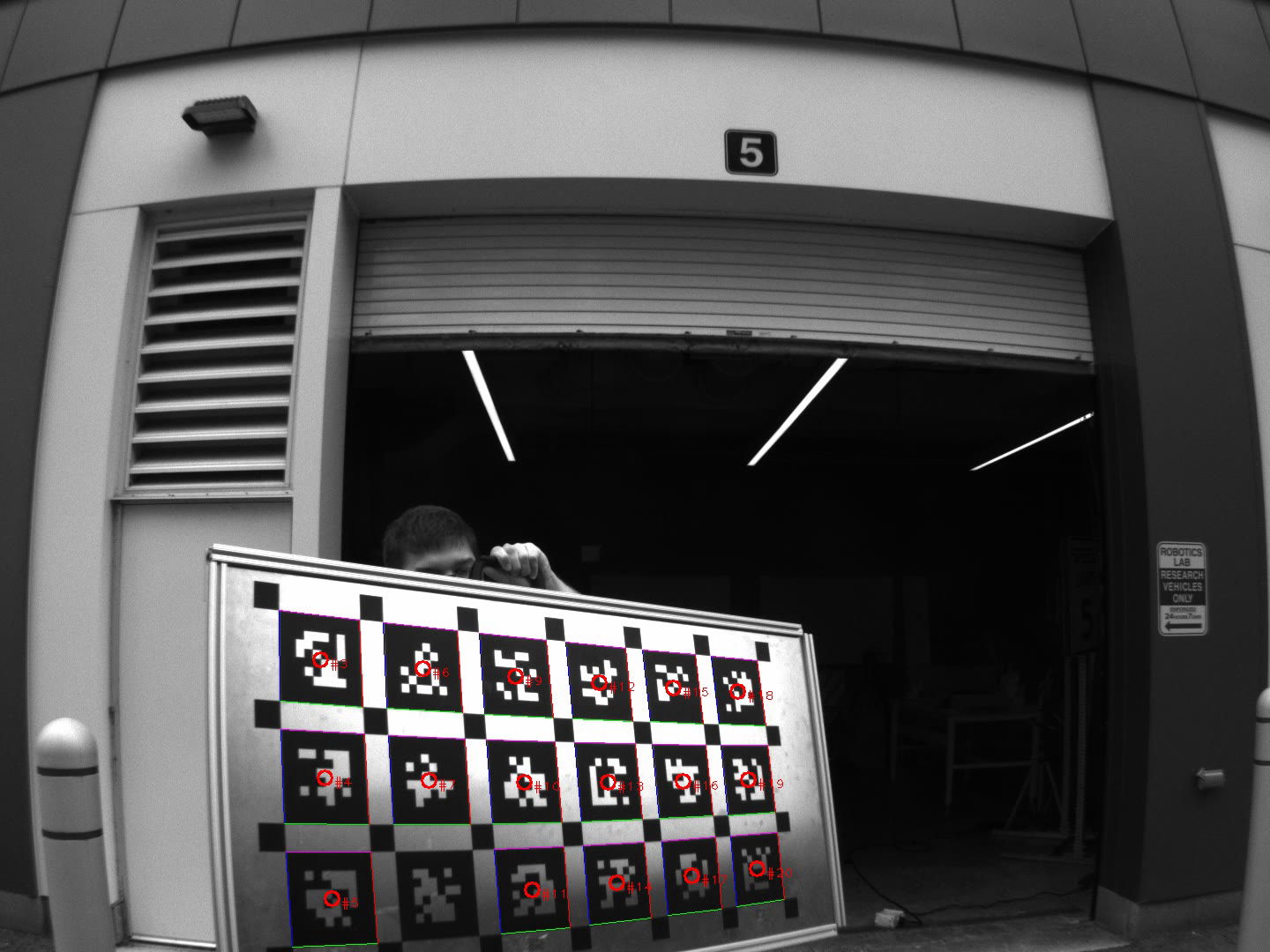}
        \caption{Monochrome image \phantom{placeholder to align subfigures}}
    \end{subfigure}
    \begin{subfigure}{0.475\columnwidth}   
        \centering 
        \includegraphics[width=\columnwidth]{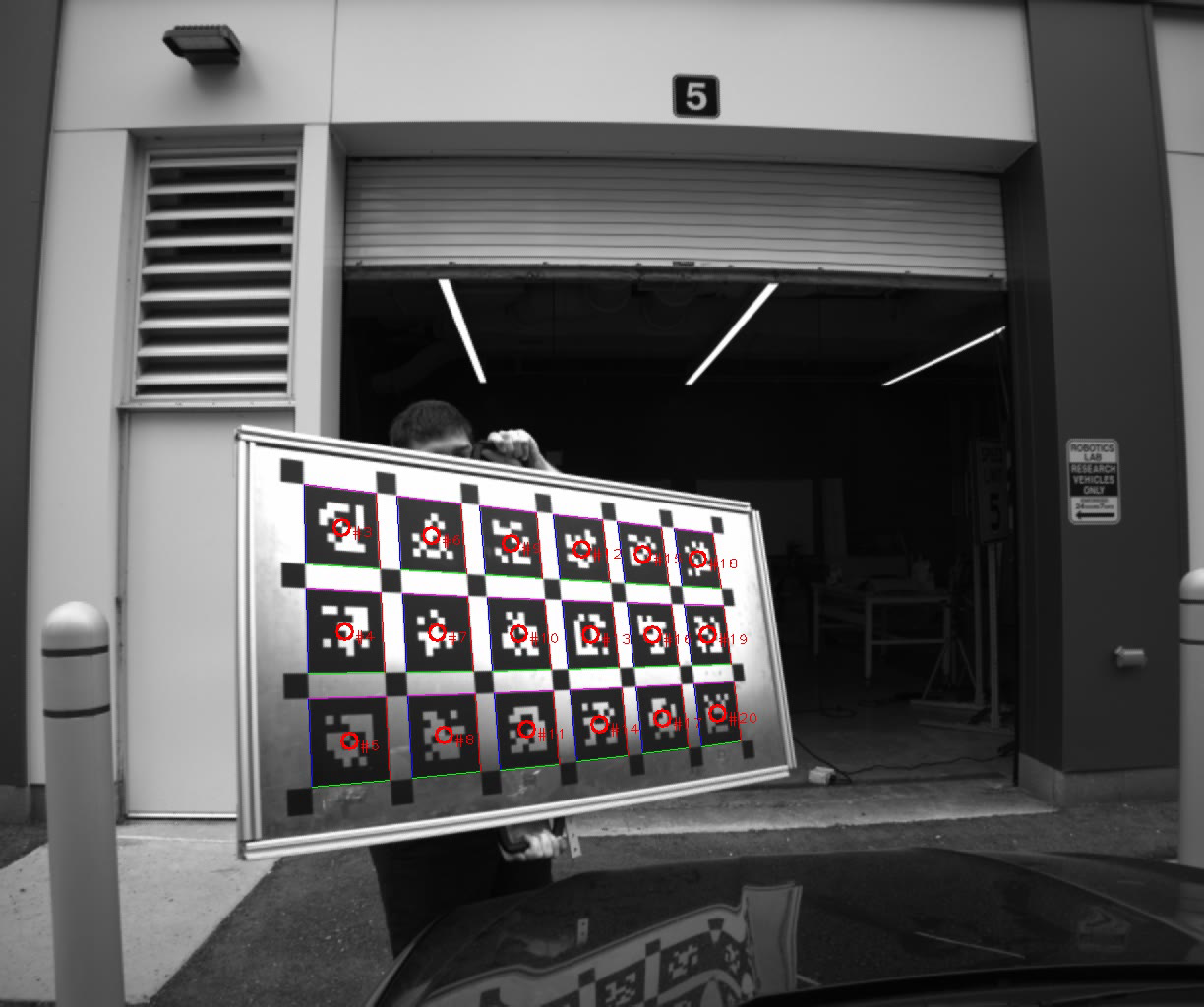}
        \caption{RGB image converted to grayscale}
    \end{subfigure}
    \caption{Calibration board AprilTag detections (denoted with red circles) in synchronized images captured or reconstructed across all modalities on the left side of the vehicle.} 
    \label{figure:calibration}
\end{figure}

We use Kalibr \cite{furgale_unified_2013} to perform simultaneous calibration of the intrinsic and extrinsic parameters across all eight cameras, applying the pinhole camera model and radial-tangential distortion model to each camera. To facilitate this multi-spectral calibration, we create a unique calibration board. Following existing work on passive thermal calibration boards \cite{elsheikh_infrared_2023}, our board is constructed with two materials of different emissivities and we heat it to increase the contrast between them. Specifically, the board is composed of an aluminum sheet with black vinyl adhered to it and cut to the shape of a grid of AprilTags \cite{olson_apriltag_2011}. The high emissivity vinyl and low emissivity aluminum primarily emit and reflect IR radiation, respectively. We apply a heat gun to the board prior to recording a calibration sequence. When heated, the vinyl appears bright in the thermal images while the aluminum reflects the generally colder surroundings and appears darker. While recording the calibration sequence, we move the board slowly to mitigate the impact of the thermal cameras' motion blur and rolling shutter effect.

The AprilTag detection employed by Kalibr can operate directly on monochrome images and RGB images converted to grayscale. To prepare the thermal images for AprilTag detection we truncate, invert, and normalize them to an 8-bit range. We use e2calib \cite{muglikar_how_2021} to reconstruct intensity images from events that are synchronized with the other modalities. Figure~\ref{figure:calibration} shows simultaneous AprilTag detections across all modalities. As noted with similar calibration boards \cite{ursine_thermal_2012, st-laurent_passive_2016}, reflections off the aluminum surface can interfere with detection. However, by using an AprilGrid rather than a checkerboard pattern \cite{huai_review_2023}, we are able to make use of partial detections of the board despite reflections and occlusions, as shown in Fig.~\ref{subfigure:calibration_thermal}. We ultimately obtain zero mean reprojection errors with standard deviations less than or equal to 0.6 pixels, except along the dominant direction of motion in the thermal camera, where the standard deviation is 1 pixel. 

\subsection{Ground Truth}\label{section:ground_truth}

We log data from the POS-LV 420 system at 200 Hz over the course of entire data collection sessions that span multiple recorded sequences. To aid IMU alignment in both forward and backward processing of the data, we begin each session by remaining static for 5 minutes and then performing dynamic movements (figure-eights and linear accelerations) and end each session with the same stages in the opposite order. We post-process the logged data with the Applanix POSPac Mobile Mapping Suite V8.8\endnote{\url{https://www.applanix.com/products/pospac-mms.htm}} using the IN-Fusion Smartbase mode \cite{hutton_tight_2023}, which employs a tightly-coupled Kalman filter to fuse data from all components of the system, along with atmospheric errors interpolated from multiple nearby base stations, to produce a highly accurate pose trajectory estimate of the vehicle base link frame relative to the earth-centered earth-fixed (ECEF) frame. The estimated standard deviations are on average 2 cm in east and north, 5 cm in height, 0.005\degree in roll and pitch, and 0.03\degree in yaw. To relate the POS-LV 420 output with the cameras we manually measure the relative pose between the vehicle base link frame and each camera frame. By using laser levels to aid in measurement, we expect the accuracy of the manually measured extrinsics to be within $\pm5$ mm and $\pm 3\degree$. The POS-LV 420 timestamps are output in GPS time, have nanosecond accuracy, and are easily converted to TAI.  

\subsection{Routes and Sequences}

\begin{table}[]
\caption{Sequences collected. The sequence names follow the pattern \texttt{$<$route$>$\_$<$direction$>$$<$lighting$>$$<$index$>$}.}
\resizebox{\columnwidth}{!}{%
\begin{tabular}{@{}ccccccc@{}}
\toprule\toprule
\textbf{\begin{tabular}[c]{@{}c@{}}Sequence\\ Name\end{tabular}} & \textbf{Route}               & \textbf{Direction}                                         & \textbf{Lighting}          & \textbf{\begin{tabular}[c]{@{}c@{}}Path\\ Length\\ (km)\end{tabular}} & \textbf{\begin{tabular}[c]{@{}c@{}}Duration\\ (minutes)\end{tabular}} & \textbf{\begin{tabular}[c]{@{}c@{}}Total\\ Size\\ (GB)\end{tabular}} \\ \midrule
\texttt{R0\_FA0}                                                 & \multirow{6}{*}{\texttt{R0}} & \multirow{3}{*}{Forward}                                   & Afternoon                  & \multirow{6}{*}{8.3}                                                  & 17.5                                                                  & 139                                                                  \\ \cmidrule(r){1-1} \cmidrule(lr){4-4} \cmidrule(l){6-7} 
\texttt{R0\_FS0}                                                 &                              &                                                            & Sunset                     &                                                                       & 15.2                                                                  & 104                                                                  \\ \cmidrule(r){1-1} \cmidrule(lr){4-4} \cmidrule(l){6-7} 
\texttt{R0\_FN0}                                                 &                              &                                                            & Night                      &                                                                       & 14.8                                                                  & 92                                                                   \\ \cmidrule(r){1-1} \cmidrule(lr){3-4} \cmidrule(l){6-7} 
\texttt{R0\_RA0}                                                 &                              & \multirow{3}{*}{Reverse}                                   & Afternoon                  &                                                                       & 21.5                                                                  & 162                                                                  \\ \cmidrule(r){1-1} \cmidrule(lr){4-4} \cmidrule(l){6-7} 
\texttt{R0\_RS0}                                                 &                              &                                                            & Sunset                     &                                                                       & 17.1                                                                  & 127                                                                  \\ \cmidrule(r){1-1} \cmidrule(lr){4-4} \cmidrule(l){6-7} 
\texttt{R0\_RN0}                                                 &                              &                                                            & Night                      &                                                                       & 15.5                                                                  & 92                                                                   \\ \midrule
\texttt{R1\_FA0}                                                 & \multirow{3}{*}{\texttt{R1}} & Forward                                                    & \multirow{3}{*}{Afternoon} & \multirow{3}{*}{8.6}                                                  & 27.5                                                                  & 217                                                                  \\ \cmidrule(r){1-1} \cmidrule(lr){3-3} \cmidrule(l){6-7} 
\texttt{R1\_DA0}                                                 &                              & \begin{tabular}[c]{@{}c@{}}Forward,\\ Diverge\end{tabular} &                            &                                                                       & 20.8                                                                  & 173                                                                  \\ \cmidrule(r){1-1} \cmidrule(lr){3-3} \cmidrule(l){6-7} 
\texttt{R1\_RA0}                                                 &                              & Reverse                                                    &                            &                                                                       & 22.4                                                                  & 169                                                                  \\ \midrule
\texttt{C1}                                                      & \multicolumn{4}{c}{Calibration}                                                                                                                                                                & 2.2                                                                   & 16                                                                   \\ \bottomrule \bottomrule
\end{tabular}%
}
\label{table:sequences}
\end{table}

\begin{figure*}
\centering
\includegraphics[width=0.93\textwidth]{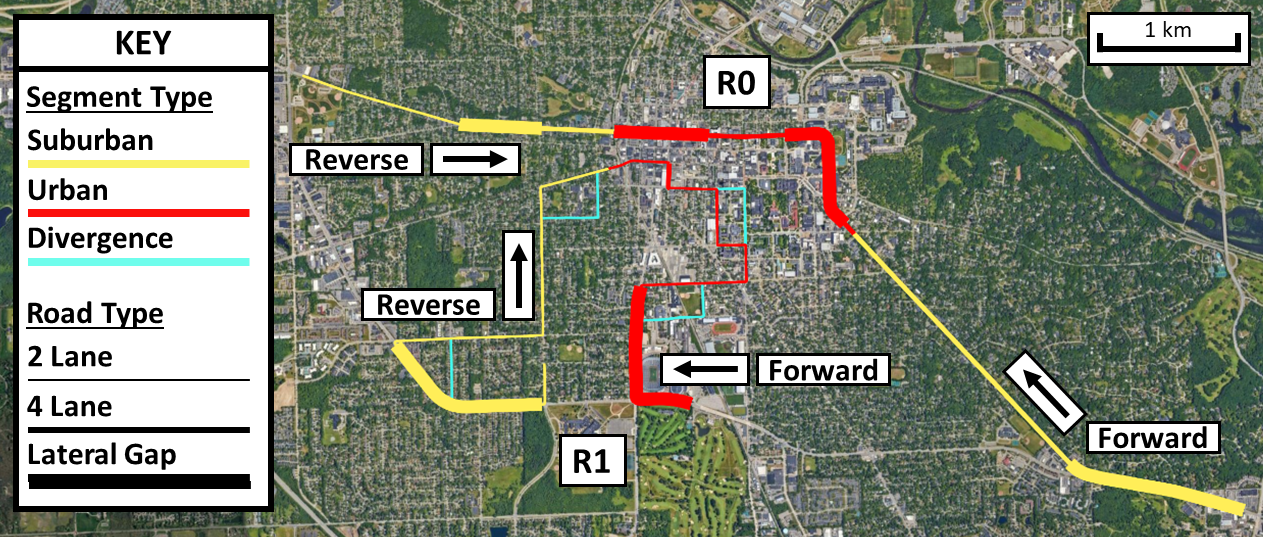}
\caption{Data collection routes R0 and R1. Scene types and divergences are distinguished with different colors. Road widths and lateral gaps between opposing lanes are distinguished with different line thicknesses. The forward and reverse labels denote the directions used in the sequence names.}  
\label{figure:routes}
\end{figure*}

Data was collected along two $\sim$8 km routes R0 and R1 as shown in Fig.~\ref{figure:routes}. The routes pass through suburban areas, characterized primarily by trees and residential buildings, and urban areas, which include large multi-story buildings. Each route was driven in both the forward and reverse directions to support opposing viewpoint place recognition evaluation. The routes cover four and two lanes roads and were mainly driven in the left lane with only the lane line dividing opposing directions. However, some portions include an additional lateral shift due to medians, turn lanes, and occasional lane shifts, which provides a further challenge for place recognition. Figure~\ref{figure:routes} shows the routes, denotes their various stages, and defines the forward and reverse directions.

Route R1 was driven in the afternoon while R0 was driven under three lighting conditions: afternoon, sunset, and night. The afternoon sequences are bright and have high thermal contrast, while the sunset and night sequences feature high dynamic range, low thermal contrast, and low light scenes. Figure~\ref{figure:examples} shows data from each sequence recorded along route R0. Route R1 was driven in the forward direction twice, with one sequence including divergences from the route to allow place recognition to be evaluated with true negatives (i.e., places in the query sequence with no match in the reference sequence). Table \ref{table:sequences} summarizes all sequences collected.

\subsection{Known Issues}

We have identified a few issues that may impact processing. We expect the overall impact to be negligible, but we mention them here for completeness. First, the auto-algorithms used by the RGB and monochrome cameras to control exposure, gain, and white balancing, run independently in each camera leading to potential differences between the left and right images of each stereo pair. However, differences in gain and exposure can be approximately corrected using the values recorded for these quantities per image as described in section~\ref{section:data_format}. Second, outside of the NUCs, 1-2 frames are dropped from the thermal cameras $\sim$10-20 times per sequence. Finally, a few insects collided with the visible camera enclosures in the route R0 sequences collected at night. These insects appear as small out-of-focus smudges at the top of each visible camera's FOV.

\section{Data Organization and Format}\label{section:data_format}

The dataset is available for download at the University of Michigan Library's Deep Blue Data repository\endnote{\url{https://deepblue.lib.umich.edu/data/collections/v118rf157}}. The data is organized by sequence. Within each sequence, the data from each sensor is provided as a separate HDF5\endnote{\url{https://www.hdfgroup.org/solutions/hdf5/}} file (abbreviated H5) and the camera calibration results (section~\ref{section:camera_calibration}) and manually measured extrinsics (section~\ref{section:ground_truth}) are given as human-readable YAML\endnote{\url{https://yaml.org/}} files. We use the H5 file format as it is guaranteed to be backward compatible, allows heterogeneous data to be stored in a single file, and supports lossless compression and efficient partial data access. The content of these files is summarized below; a more detailed description is available in our software tools repository (section \ref{section:software}).

The camera data filenames follow the pattern \texttt{$<$sequence name$>$\_$<$sensor$>$\_$<$side$>$.h5} where \texttt{$<$side$>$} is \texttt{left} or \texttt{right} and \texttt{$<$sensor$>$} denotes the camera type or model as \texttt{mono} (monochrome), \texttt{rgb} (RGB), \texttt{adk} (thermal), or \texttt{dvxplorer} (event). Each of the frame-based camera files include a H5 group, \texttt{image\_raw}, containing raw images and their corresponding timestamps as H5 datasets \texttt{images} and \texttt{timestamps}. Similarly, the event camera files include a H5 group, \texttt{events}, which contains a separate H5 dataset for each event field: \texttt{x\_coordinates}, \texttt{y\_coordinates}, \texttt{timestamps}, and \texttt{polarities}. Fixed quantities relating to the images and events, such as width, height, and encoding, are attached as H5 attributes to the \texttt{image\_raw} and \texttt{events} H5 groups. Additionally, the frame-based camera files include an \texttt{image\_meta} H5 group containing per-image metadata such as exposure, gain, or NUC status. Comprehensive camera settings are also attached as attributes to the root H5 group in the monochrome, RGB, and event camera files.

The ground truth poses are provided in a file named \texttt{$<$sequence name$>$\_applanix.h5}. This file includes a \texttt{pose\_base\_link} H5 group containing H5 datasets \texttt{positions}, \texttt{quaternions}, and \texttt{timestamps}, which give the poses of the base link frame in the ECEF frame.

\section{Software Tools}\label{section:software}

To facilitate efficient use of the dataset we provide a set of software tools\endnote{\url{https://github.com/umautobots/nsavp_tools}}. To ensure cross-platform compatibility, we include Dockerfiles alongside instructions to build Docker images capable of running all provided code.

\begin{figure}
\centering
\includegraphics[width=1.0\columnwidth]{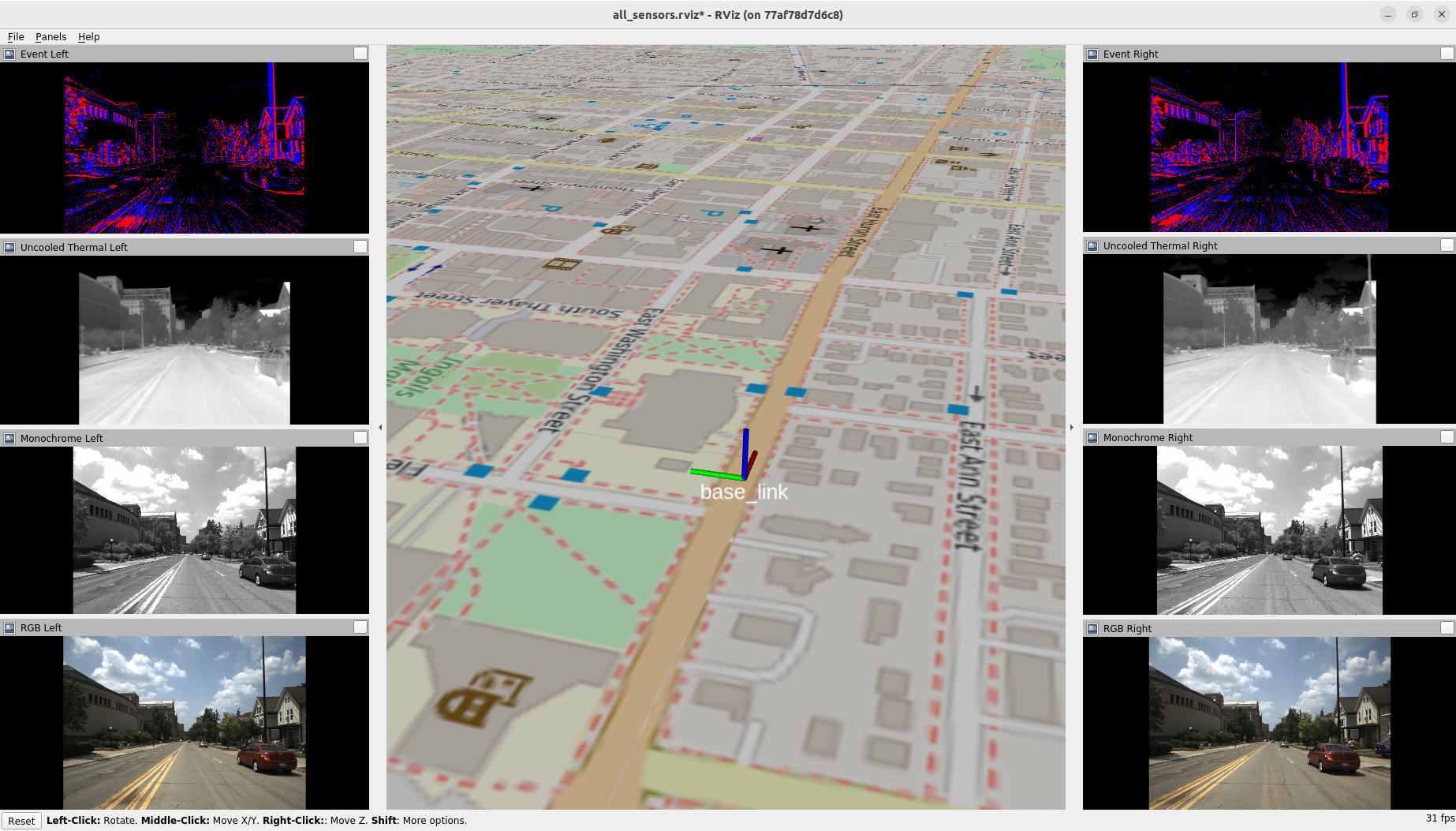}
\caption[RViz visualization of ground truth pose and pre-processed camera data (map sourced from OpenStreetMap).]{RViz visualization of ground truth pose and pre-processed camera data (map sourced from OpenStreetMap\endnote{\url{https://www.openstreetmap.org}}).}  
\label{figure:visualization}
\end{figure}

The \texttt{conversion} ROS package includes scripts to combine H5 files and convert a combined H5 file to a rosbag or to directories containing human readable CSV files and PNG images. Reading the data from the rosbag format, the \texttt{preprocessing\_and\_visualization} ROS package includes scripts that demonstrate common pre-processing operations (debayering, like and cross modality rectification, cropping, etc.) and visualize the camera data and ground truth pose as shown in Fig.~\ref{figure:visualization}.

\begin{figure}
\centering
\includegraphics[width=1.0\columnwidth]{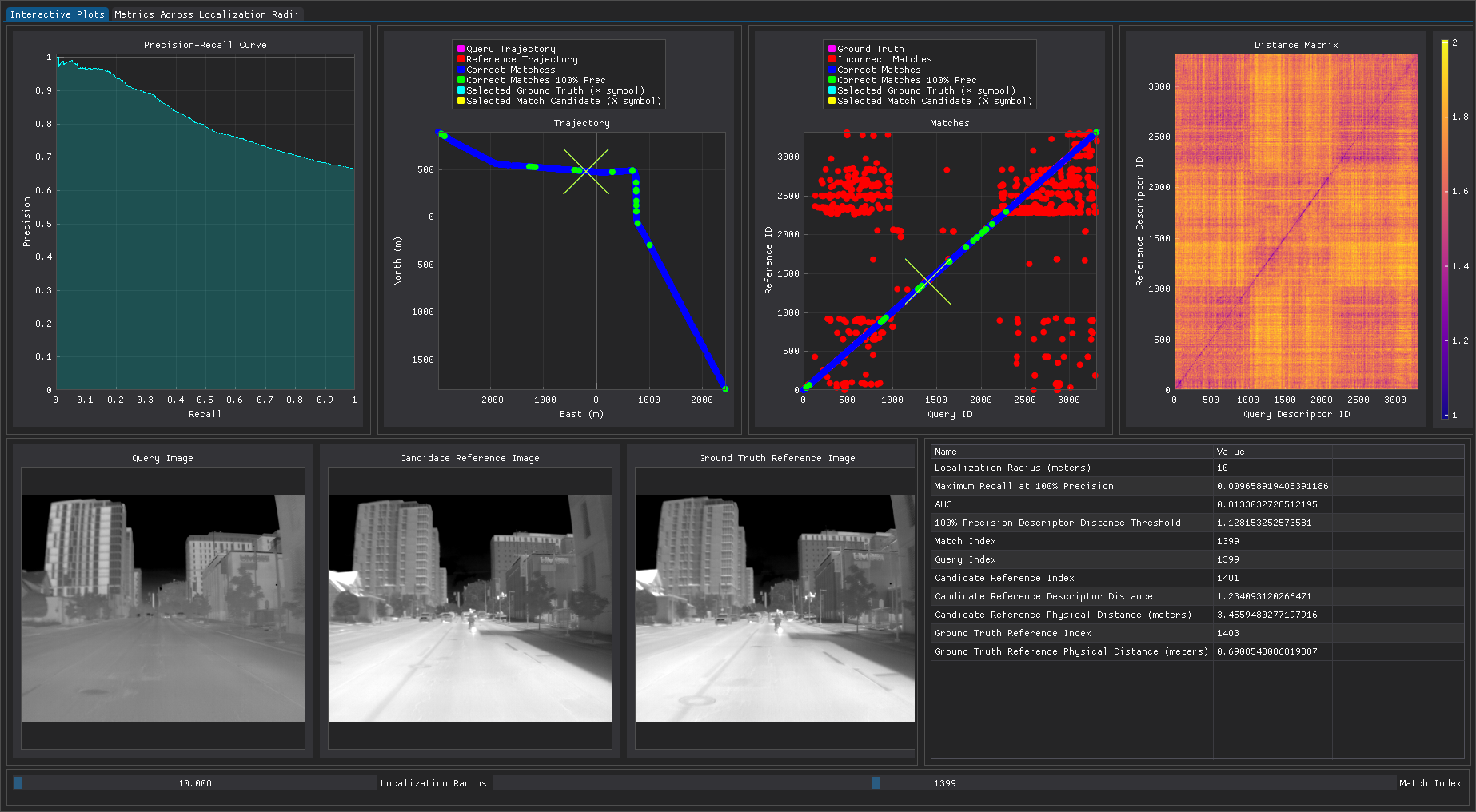}
\caption{Place recognition evaluation GUI including the following components from left-to-right and top-to-bottom: precision-recall curve, plot of correct matches along the trajectory, plot of candidate and ground truth matches, distance matrix, query image, candidate reference image, ground truth reference image, table of place recognition metrics, localization radius slider, and match index slider.} 
\label{figure:place_recognition_gui}
\end{figure}

Additionally, we include example scripts applying and evaluating a place recognition algorithm with the dataset, utilizing the H5 format directly. The example uses NetVLAD \cite{arandjelovic_netvlad_2016} and produces the same results given for this algorithm in section~\ref{section:benchmark}. As part of this example, we also provide an algorithm-agnostic GUI that can be used to visualize place recognition results and step through individual matches. Figure~\ref{figure:place_recognition_gui} shows the GUI displaying the results of running NetVLAD with the left thermal images from the \texttt{R0\_RA0} sequence and \texttt{R0\_RN0} sequence as described in the following section. 

\section{Place Recognition Benchmark}\label{section:benchmark}

To highlight some of the research challenges our dataset poses, we demonstrate a use case of our dataset on the task of place recognition. Place recognition is the problem of determining whether a given place, represented with a query, matches a specific place within a database of references. We evaluate the opposing viewpoint focused method LoST-X \cite{garg_lost_2018} and the seminal method NetVLAD \cite{arandjelovic_netvlad_2016}. As in \cite{garg_lost_2018}, we select input images to be 2 meters apart according to the ground truth position data, and additionally augment LoST-X and NetVLAD with sequence matching \cite{milford_seqslam_2012} using a sequence length of 51 (denoted with +SM). We apply both approaches to the left RGB images of each selected sequence and also run NetVLAD with the left thermal images (denoted with +T). The thermal images are truncated and converted to 8-bit at the input to NetVLAD.

In all tests, the reference database is formed using a 6.7 km subset of the \texttt{R0\_RA0} sequence. The queries are taken from the \texttt{R0\_RS0}, \texttt{R0\_RN0}, \texttt{R0\_FA0}, and \texttt{R0\_FN0} sequences using the same subset. For succinctness, the sequence pairs will be referred to by the sequence the queries are drawn from. We use a 10 meter localization radius (denoted with $r=10$ m) to determine correct matches for the sequences driven in the same direction as the reference database and, following \cite{garg_lost_2018}, we use an 80 meter localization radius (denoted with $r=80$ m) for the opposing viewpoint sequences.

\begin{figure}
\centering
\includegraphics[width=1.0\columnwidth]{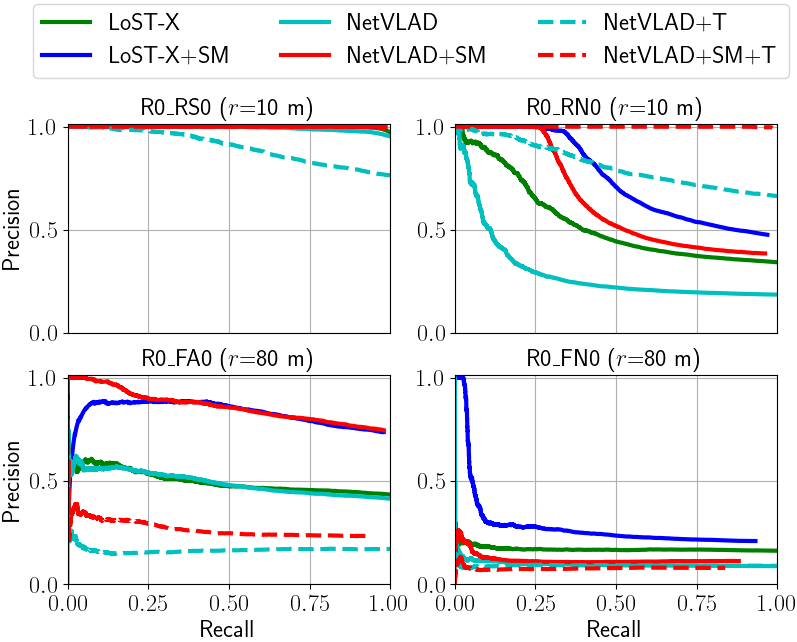}
\caption{Precision-recall curves of each method on the selected sequence pairs.} 
\label{figure:benchmark}
\end{figure}

Figure~\ref{figure:benchmark} shows precision-recall curves of each tested method on the selected sequence pairs. All methods perform well with the \texttt{R0\_RSO} sequence, which was driven in the same direction and under similar lighting conditions as the reference database. The \texttt{R0\_RN0} sequence is more difficult, due to the drastic lighting change and poor visibility in the RGB camera. This demonstrates a common failure case for visible spectrum cameras (RGB and monochrome) deployed for AV perception. This issue can be overcome by using thermal images instead, as shown by the NetVLAD+SM+T result. Still, the opposing viewpoint sequences, \texttt{R0\_FA0} and \texttt{R0\_FN0}, provide a further challenge and the limited results on these sequences indicate there is significant room for further development with both visible spectrum and thermal cameras. The NSAVP dataset provides a resource for the research community to investigate this open challenge.

\section{Summary and Future Work}

We introduce the NSAVP dataset, which features stereo event, thermal, monochrome, and RGB cameras and captures opposing viewpoints and varied lighting conditions. We believe the dataset presents unique opportunities to apply novel sensors to AV perception. We demonstrate an example use case of this dataset towards the task of place recognition. In the near future, we aim to expand the sensor suite to include lidar and IMU sensors and additionally collect data in adverse weather conditions including snow and rain, which further challenge conventional sensors.

\begin{acks}
The authors wish to thank Martin Ward for his work in prototyping the hardware design, and Rahul Agrawal, Vivek Jaiswal, Dhiraj Maji, and Xun Tu for their contributions to the data collection process and software.
\end{acks}

\begin{dci}
The Authors declare that there is no conflict of interest
\end{dci}

\begin{funding}
The authors disclosed receipt of the following financial support for the research, authorship, and/or publication of this article: This work was supported by the Ford Motor Company via the Ford-UM Alliance under award N028603.
\end{funding}

\theendnotes

\bibliographystyle{SageH}
\bibliography{references.bib}

\begin{thebibliography}{54}
\providecommand{\natexlab}[1]{#1}
\providecommand{\url}[1]{\texttt{#1}}
\providecommand{\urlprefix}{URL }
\expandafter\ifx\csname urlstyle\endcsname\relax
  \providecommand{\doi}[1]{DOI:\discretionary{}{}{}#1}\else
  \providecommand{\doi}{DOI:\discretionary{}{}{}\begingroup
  \urlstyle{rm}\Url}\fi

\bibitem[{Agarwal et~al.(2020)Agarwal, Vora, Pandey, Williams, Kourous and
  McBride}]{ford_multi_av}
Agarwal S, Vora A, Pandey G, Williams W, Kourous H and McBride J (2020) Ford
  multi-{AV} seasonal dataset.
\newblock \emph{The International Journal of Robotics Research} 39(12):
  1367--1376.

\bibitem[{Alonso and Murillo(2019)}]{alonso_ev-segnet_2019}
Alonso I and Murillo AC (2019) {EV}-{SegNet}: Semantic segmentation for
  event-based cameras.
\newblock In: \emph{2019 IEEE/CVF Conference on Computer Vision and Pattern
  Recognition Workshops (CVPRW)}. Long Beach, CA, USA: IEEE, pp. 1624--1633.

\bibitem[{Arandjelovic et~al.(2016)Arandjelovic, Gronat, Torii, Pajdla and
  Sivic}]{arandjelovic_netvlad_2016}
Arandjelovic R, Gronat P, Torii A, Pajdla T and Sivic J (2016) {NetVLAD}: {CNN}
  architecture for weakly supervised place recognition.
\newblock In: \emph{2016 IEEE Conference on Computer Vision and Pattern
  Recognition (CVPR)}. Las Vegas, NV, USA: IEEE, pp. 5297--5307.

\bibitem[{Binas et~al.(2017)Binas, Neil, Liu and Delbruck}]{binas_ddd17_2017}
Binas J, Neil D, Liu SC and Delbruck T (2017) {DDD17}: End-to-end {DAVIS}
  driving dataset.
\newblock \emph{arXiv preprint arXiv:1711.01458} .

\bibitem[{Chaney et~al.(2023)Chaney, Cladera, Wang, Bisulco, Hsieh, Korpela,
  Kumar, Taylor and Daniilidis}]{chaney_m3ed_2023}
Chaney K, Cladera F, Wang Z, Bisulco A, Hsieh MA, Korpela C, Kumar V, Taylor CJ
  and Daniilidis K (2023) {M3ED}: Multi-robot, multi-sensor, multi-environment
  event dataset.
\newblock In: \emph{2023 IEEE/CVF Conference on Computer Vision and Pattern
  Recognition Workshops (CVPRW)}. Vancouver, BC, Canada: IEEE, pp. 4016--4023.

\bibitem[{Chen et~al.(2023)Chen, Guan, Huang, Zhong, Wen, Hsu and
  Lu}]{chen_ecmd_2023}
Chen P, Guan W, Huang F, Zhong Y, Wen W, Hsu LT and Lu P (2023) {ECMD}: An
  event-centric multisensory driving dataset for {SLAM}.
\newblock \emph{IEEE Transactions on Intelligent Vehicles} : 1--10.

\bibitem[{Cheng et~al.(2019)Cheng, Luo, Yang, Yu, Chen and Li}]{cheng_det_2019}
Cheng W, Luo H, Yang W, Yu L, Chen S and Li W (2019) {DET}: A high-resolution
  {DVS} dataset for lane extraction.
\newblock In: \emph{2019 IEEE/CVF Conference on Computer Vision and Pattern
  Recognition Workshops (CVPRW)}. Long Beach, CA, USA: IEEE, pp. 1666--1675.

\bibitem[{Choi et~al.(2018)Choi, Kim, Hwang, Park, Yoon, An and
  Kweon}]{choi_kaist_2018}
Choi Y, Kim N, Hwang S, Park K, Yoon JS, An K and Kweon IS (2018) {KAIST}
  multi-spectral day/night data set for autonomous and assisted driving.
\newblock \emph{IEEE Transactions on Intelligent Transportation Systems} 19(3):
  934--948.

\bibitem[{De~Tournemire et~al.(2020)De~Tournemire, Nitti, Perot, Migliore and
  Sironi}]{de_tournemire_large_2020}
De~Tournemire P, Nitti D, Perot E, Migliore D and Sironi A (2020) A large scale
  event-based detection dataset for automotive.
\newblock \emph{arXiv preprint arXiv:2001.08499} .

\bibitem[{ElSheikh et~al.(2023)ElSheikh, Abu-Nabah, Hamdan and
  Tian}]{elsheikh_infrared_2023}
ElSheikh A, Abu-Nabah BA, Hamdan MO and Tian GY (2023) Infrared camera
  geometric calibration: A review and a precise thermal radiation checkerboard
  target.
\newblock \emph{Sensors} 23(7): 3479.

\bibitem[{Farooq et~al.(2023)Farooq, Shariff and
  Corcoran}]{farooq_evaluation_2023}
Farooq MA, Shariff W and Corcoran P (2023) Evaluation of thermal imaging on
  embedded {GPU} platforms for application in vehicular assistance systems.
\newblock \emph{IEEE Transactions on Intelligent Vehicles} 8(2): 1130--1144.

\bibitem[{Fischer and Milford(2020)}]{fischer_event-based_2020}
Fischer T and Milford M (2020) Event-based visual place recognition with
  ensembles of temporal windows.
\newblock \emph{IEEE Robotics and Automation Letters} 5(4): 6924--6931.

\bibitem[{Franchi et~al.(2024)Franchi, Hariat, Yu, Belkhir, Manzanera and
  Filliat}]{franchi_infraparis_2023}
Franchi G, Hariat M, Yu X, Belkhir N, Manzanera A and Filliat D (2024)
  {InfraParis}: A multi-modal and multi-task autonomous driving dataset.
\newblock In: \emph{Proceedings of the IEEE/CVF Winter Conference on
  Applications of Computer Vision (WACV)}. Waikoloa, HI, USA: IEEE, pp.
  2973--2983.

\bibitem[{Furgale et~al.(2013)Furgale, Rehder and
  Siegwart}]{furgale_unified_2013}
Furgale P, Rehder J and Siegwart R (2013) Unified temporal and spatial
  calibration for multi-sensor systems.
\newblock In: \emph{2013 IEEE/RSJ International Conference on Intelligent
  Robots and Systems}. Tokyo, Japan: IEEE, pp. 1280--1286.

\bibitem[{Gallego et~al.(2022)Gallego, Delbruck, Orchard, Bartolozzi, Taba,
  Censi, Leutenegger, Davison, Conradt, Daniilidis and
  Scaramuzza}]{gallego_event-based_2022}
Gallego G, Delbruck T, Orchard G, Bartolozzi C, Taba B, Censi A, Leutenegger S,
  Davison AJ, Conradt J, Daniilidis K and Scaramuzza D (2022) Event-based
  vision: A survey.
\newblock \emph{IEEE Transactions on Pattern Analysis and Machine Intelligence}
  44(1): 154--180.

\bibitem[{Garg et~al.(2018)Garg, Suenderhauf and Milford}]{garg_lost_2018}
Garg S, Suenderhauf N and Milford M (2018) {LoST}? {Appearance}-invariant place
  recognition for opposite viewpoints using visual semantics.
\newblock In: \emph{Proceedings of Robotics: Science and Systems}. Pittsburgh,
  Pennsylvania: Robotics: Science and Systems Foundation.

\bibitem[{Gehrig et~al.(2021)Gehrig, Aarents, Gehrig and
  Scaramuzza}]{gehrig_dsec_2021}
Gehrig M, Aarents W, Gehrig D and Scaramuzza D (2021) {DSEC}: A stereo event
  camera dataset for driving scenarios.
\newblock \emph{IEEE Robotics and Automation Letters} 6(3): 4947--4954.

\bibitem[{Geiger et~al.(2013)Geiger, Lenz, Stiller and
  Urtasun}]{geiger_vision_2013}
Geiger A, Lenz P, Stiller C and Urtasun R (2013) Vision meets robotics: The
  {KITTI} dataset.
\newblock \emph{The International Journal of Robotics Research} 32(11):
  1231--1237.

\bibitem[{Ha et~al.(2017)Ha, Watanabe, Karasawa, Ushiku and
  Harada}]{ha_mfnet_2017}
Ha Q, Watanabe K, Karasawa T, Ushiku Y and Harada T (2017) {MFNet}: Towards
  real-time semantic segmentation for autonomous vehicles with multi-spectral
  scenes.
\newblock In: \emph{2017 {IEEE}/{RSJ} {International} {Conference} on
  {Intelligent} {Robots} and {Systems} ({IROS})}. Vancouver, BC, Canada: IEEE,
  pp. 5108--5115.

\bibitem[{Hadviger et~al.(2021)Hadviger, Cvišić, Marković, Vražić and
  Petrović}]{hadviger_feature-based_2021}
Hadviger A, Cvišić I, Marković I, Vražić S and Petrović I (2021)
  Feature-based event stereo visual odometry.
\newblock In: \emph{2021 European Conference on Mobile Robots (ECMR)}. Bonn,
  Germany: IEEE, pp. 1--6.

\bibitem[{Hadviger et~al.(2023)Hadviger, Štironja, Cvišić, Marković,
  Vražić and Petrović}]{hadviger_stereo_2023}
Hadviger A, Štironja VJ, Cvišić I, Marković I, Vražić S and Petrović I
  (2023) Stereo visual localization dataset featuring event cameras.
\newblock In: \emph{2023 European Conference on Mobile Robots (ECMR)}. Coimbra,
  Portugal: IEEE, pp. 1--6.

\bibitem[{Hu et~al.(2020)Hu, Binas, Neil, Liu and Delbruck}]{hu_ddd20_2020}
Hu Y, Binas J, Neil D, Liu SC and Delbruck T (2020) {DDD20} end-to-end event
  camera driving dataset: Fusing frames and events with deep learning for
  improved steering prediction.
\newblock In: \emph{2020 IEEE 23rd International Conference on Intelligent
  Transportation Systems (ITSC)}. Rhodes, Greece: IEEE, pp. 1--6.

\bibitem[{Hu et~al.(2021)Hu, Liu and Delbruck}]{hu_v2e_2021}
Hu Y, Liu SC and Delbruck T (2021) {v2e}: From video frames to realistic {DVS}
  events.
\newblock In: \emph{2021 IEEE/CVF Conference on Computer Vision and Pattern
  Recognition Workshops (CVPRW)}. Nashville, TN, USA: IEEE, pp. 1312--1321.

\bibitem[{Huai et~al.(2023)Huai, Zhuang, Shao, Jozkow, Wang, Liu, He and
  Yilmaz}]{huai_review_2023}
Huai J, Zhuang Y, Shao Y, Jozkow G, Wang B, Liu J, He Y and Yilmaz A (2023) A
  review and comparative study of close-range geometric camera calibration
  tools.
\newblock \emph{arXiv preprint arXiv:2306.09014} .

\bibitem[{Hutton et~al.(2008)Hutton, Ip, Bourke, Scherzinger, Gopaul, Canter,
  Oveland and Blankenberg}]{hutton_tight_2023}
Hutton J, Ip A, Bourke T, Scherzinger B, Gopaul N, Canter P, Oveland I and
  Blankenberg L (2008) Tight integration of {GNSS} post-processed virtual
  reference station with inertial data for increased accuracy and productivity
  of airborne mapping.
\newblock \emph{Remote Sensing and Spatial Information Sciences} 37(B5):
  829--834.

\bibitem[{Hwang et~al.(2015)Hwang, Park, Kim, Choi and
  Kweon}]{hwang_multispectral_2015}
Hwang S, Park J, Kim N, Choi Y and Kweon IS (2015) Multispectral pedestrian
  detection: Benchmark dataset and baseline.
\newblock In: \emph{2015 IEEE Conference on Computer Vision and Pattern
  Recognition (CVPR)}. Boston, MA, USA: IEEE, pp. 1037--1045.

\bibitem[{Judd et~al.(2019)Judd, Thornton and Richards}]{judd_automotive_2019}
Judd KM, Thornton MP and Richards AA (2019) Automotive sensing: assessing the
  impact of fog on {LWIR}, {MWIR}, {SWIR}, visible, and lidar performance.
\newblock In: \emph{Infrared {Technology} and {Applications} {XLV}}. Baltimore,
  MD, USA: SPIE, pp. 322--334.

\bibitem[{Lee et~al.(2022)Lee, Cho, Shin, Kim and Myung}]{lee_vivid_2022}
Lee AJ, Cho Y, Shin Ys, Kim A and Myung H (2022) {ViViD}++ : Vision for
  visibility dataset.
\newblock \emph{IEEE Robotics and Automation Letters} 7(3): 6282--6289.

\bibitem[{Ligocki et~al.(2020)Ligocki, Jelinek and Zalud}]{ligocki_brno_2020}
Ligocki A, Jelinek A and Zalud L (2020) Brno urban dataset - the new data for
  self-driving agents and mapping tasks.
\newblock In: \emph{2020 IEEE International Conference on Robotics and
  Automation (ICRA)}. Paris, France: IEEE, pp. 3284--3290.

\bibitem[{Matthies and Shafer(1987)}]{matthies_error_1987}
Matthies L and Shafer S (1987) Error modeling in stereo navigation.
\newblock \emph{IEEE Journal on Robotics and Automation} 3(3): 239--248.

\bibitem[{Milford and Wyeth(2012)}]{milford_seqslam_2012}
Milford MJ and Wyeth GF (2012) {SeqSLAM}: Visual route-based navigation for
  sunny summer days and stormy winter nights.
\newblock In: \emph{2012 IEEE International Conference on Robotics and
  Automation}. Saint Paul, MN, USA: IEEE, pp. 1643--1649.

\bibitem[{Miron et~al.(2015)Miron, Rogozan, Ainouz, Bensrhair and
  Broggi}]{miron_evaluation_2015}
Miron A, Rogozan A, Ainouz S, Bensrhair A and Broggi A (2015) An evaluation of
  the pedestrian classification in a multi-domain multi-modality setup.
\newblock \emph{Sensors} 15(6): 13851--13873.

\bibitem[{Mollica et~al.(2023)Mollica, Felicioni, Legittimo, Meli, Costante and
  Valigi}]{mollica_ma-vied_2023}
Mollica G, Felicioni S, Legittimo M, Meli L, Costante G and Valigi P (2023)
  {MA}-{VIED}: A multisensor automotive visual inertial event dataset.
\newblock \emph{IEEE Transactions on Intelligent Transportation Systems} :
  1--11.

\bibitem[{Muglikar et~al.(2021)Muglikar, Gehrig, Gehrig and
  Scaramuzza}]{muglikar_how_2021}
Muglikar M, Gehrig M, Gehrig D and Scaramuzza D (2021) How to calibrate your
  event camera.
\newblock In: \emph{2021 IEEE/CVF Conference on Computer Vision and Pattern
  Recognition Workshops (CVPRW)}. Nashville, TN, USA: IEEE, pp. 1403--1409.

\bibitem[{Olson(2011)}]{olson_apriltag_2011}
Olson E (2011) {AprilTag}: A robust and flexible visual fiducial system.
\newblock In: \emph{2011 IEEE International Conference on Robotics and
  Automation}. Shanghai, China: IEEE, pp. 3400--3407.

\bibitem[{Perot et~al.(2020)Perot, de~Tournemire, Nitti, Masci and
  Sironi}]{perot_learning_2020}
Perot E, de~Tournemire P, Nitti D, Masci J and Sironi A (2020) Learning to
  detect objects with a 1 megapixel event camera.
\newblock In: \emph{Advances in Neural Information Processing Systems},
  volume~33. Curran Associates, Inc., pp. 16639--16652.

\bibitem[{Rebecq et~al.(2021)Rebecq, Ranftl, Koltun and
  Scaramuzza}]{rebecq_high_2021}
Rebecq H, Ranftl R, Koltun V and Scaramuzza D (2021) High speed and high
  dynamic range video with an event camera.
\newblock \emph{IEEE Transactions on Pattern Analysis and Machine Intelligence}
  43(6): 1964--1980.

\bibitem[{Scheerlinck et~al.(2019)Scheerlinck, Rebecq, Stoffregen, Barnes,
  Mahony and Scaramuzza}]{scheerlinck_ced_2019}
Scheerlinck C, Rebecq H, Stoffregen T, Barnes N, Mahony R and Scaramuzza D
  (2019) {CED}: Color event camera dataset.
\newblock In: \emph{2019 IEEE/CVF Conference on Computer Vision and Pattern
  Recognition Workshops (CVPRW)}. Long Beach, CA, USA: IEEE, pp. 1684--1693.

\bibitem[{Shin et~al.(2023)Shin, Park and Kweon}]{shin_deep_2023}
Shin U, Park J and Kweon IS (2023) Deep depth estimation from thermal image.
\newblock In: \emph{2023 IEEE/CVF Conference on Computer Vision and Pattern
  Recognition (CVPR)}. Vancouver, BC, Canada: IEEE, pp. 1043--1053.

\bibitem[{Sironi et~al.(2018)Sironi, Brambilla, Bourdis, Lagorce and
  Benosman}]{sironi_hats_2018}
Sironi A, Brambilla M, Bourdis N, Lagorce X and Benosman R (2018) {HATS}:
  Histograms of averaged time surfaces for robust event-based object
  classification.
\newblock In: \emph{2018 IEEE/CVF Conference on Computer Vision and Pattern
  Recognition}. Salt Lake City, UT, USA: IEEE, pp. 1731--1740.

\bibitem[{St-Laurent et~al.(2017)St-Laurent, Mikhnevich, Bubel and
  Pr{\'e}vost}]{st-laurent_passive_2016}
St-Laurent L, Mikhnevich M, Bubel A and Pr{\'e}vost D (2017) Passive
  calibration board for alignment of {VIS}-{NIR}, {SWIR} and {LWIR} images.
\newblock \emph{Quantitative InfraRed Thermography Journal} 14(2): 193--205.

\bibitem[{Sun et~al.(2022)Sun, Messikommer, Gehrig and
  Scaramuzza}]{avidan_ess_2022}
Sun Z, Messikommer N, Gehrig D and Scaramuzza D (2022) {ESS}: Learning
  event-based semantic segmentation from still images.
\newblock In: \emph{European Conference on Computer Vision}. Tel Aviv, Israel:
  Springer, pp. 341--357.

\bibitem[{Takumi et~al.(2017)Takumi, Watanabe, Ha, Tejero-De-Pablos, Ushiku and
  Harada}]{takumi_multispectral_2017}
Takumi K, Watanabe K, Ha Q, Tejero-De-Pablos A, Ushiku Y and Harada T (2017)
  Multispectral object detection for autonomous vehicles.
\newblock In: \emph{Proceedings of the on {Thematic} {Workshops} of {ACM}
  {Multimedia}}. Mountain View, CA, USA: ACM, pp. 35--43.

\bibitem[{{Teledyne FLIR}(2018)}]{noauthor_free_nodate}
{Teledyne FLIR} (2018) Teledyne {FLIR} thermal dataset for algorithm training.
\newblock Available at: \url{https://www.flir.com/oem/adas/adas-dataset-form/}
  (accessed 5 December 2023).

\bibitem[{Torres and Hayat(2003)}]{torres_kalman_2003}
Torres SN and Hayat MM (2003) Kalman filtering for adaptive nonuniformity
  correction in infrared focal-plane arrays.
\newblock \emph{Journal of the Optical Society of America A} 20(3): 470--480.

\bibitem[{Tumas et~al.(2020)Tumas, Nowosielski and
  Serackis}]{tumas_pedestrian_2020}
Tumas P, Nowosielski A and Serackis A (2020) Pedestrian detection in severe
  weather conditions.
\newblock \emph{IEEE Access} 8: 62775--62784.

\bibitem[{Ursine et~al.(2012)Ursine, Calado, Teixeira, Diniz, Silvino and
  De~Andrade}]{ursine_thermal_2012}
Ursine W, Calado F, Teixeira G, Diniz H, Silvino S and De~Andrade R (2012)
  Thermal/visible autonomous stereo visio system calibration methodology for
  non-controlled environments.
\newblock In: \emph{11th International Conference on Quantitative Infrared
  Thermography}. Naples, Italy, pp. 1--10.

\bibitem[{Valverde et~al.(2021)Valverde, Hurtado and
  Valada}]{valverde_there_2021}
Valverde FR, Hurtado JV and Valada A (2021) There is more than meets the eye:
  Self-supervised multi-object detection and tracking with sound by distilling
  multimodal knowledge.
\newblock In: \emph{2021 IEEE/CVF Conference on Computer Vision and Pattern
  Recognition (CVPR)}. Nashville, TN, USA: IEEE, pp. 11612--11621.

\bibitem[{Vertens et~al.(2020)Vertens, Zürn and
  Burgard}]{vertens_heatnet_2020}
Vertens J, Zürn J and Burgard W (2020) {HeatNet}: Bridging the day-night
  domain gap in semantic segmentation with thermal images.
\newblock In: \emph{2020 IEEE/RSJ International Conference on Intelligent
  Robots and Systems (IROS)}. Las Vegas, NV, USA: IEEE, pp. 8461--8468.

\bibitem[{Xu et~al.(2019)Xu, Zhuang, Liu, Zhou and Peng}]{XU2019199}
Xu Z, Zhuang J, Liu Q, Zhou J and Peng S (2019) Benchmarking a large-scale
  {FIR} dataset for on-road pedestrian detection.
\newblock \emph{Infrared Physics \& Technology} 96: 199--208.

\bibitem[{Yun et~al.(2022)Yun, Jung, Kim, Jung, Cho, Jeon, Kim and
  Kim}]{yun_sthereo_2022}
Yun S, Jung M, Kim J, Jung S, Cho Y, Jeon MH, Kim G and Kim A (2022) {STheReO}:
  Stereo thermal dataset for research in odometry and mapping.
\newblock In: \emph{2022 IEEE/RSJ International Conference on Intelligent
  Robots and Systems (IROS)}. Kyoto, Japan: IEEE, pp. 3857--3864.

\bibitem[{Zhang et~al.(2023{\natexlab{a}})Zhang, Zhuge, Liu, Peng, Wu, Zhang,
  Lyu, Li, Zhao, Kircali, Mharolkar, Yang, Yi, Wang and
  Wang}]{zhang_ntu4dradlm_2023}
Zhang J, Zhuge H, Liu Y, Peng G, Wu Z, Zhang H, Lyu Q, Li H, Zhao C, Kircali D,
  Mharolkar S, Yang X, Yi S, Wang Y and Wang D (2023{\natexlab{a}})
  {NTU4DRadLM}: {4D} radar-centric multi-modal dataset for localization and
  mapping.
\newblock \emph{arXiv preprint arXiv:2309.00962} .

\bibitem[{Zhang et~al.(2023{\natexlab{b}})Zhang, Carballo, Yang and
  Takeda}]{zhang_perception_2023}
Zhang Y, Carballo A, Yang H and Takeda K (2023{\natexlab{b}}) Perception and
  sensing for autonomous vehicles under adverse weather conditions: {A} survey.
\newblock \emph{ISPRS Journal of Photogrammetry and Remote Sensing} 196:
  146--177.

\bibitem[{Zhu et~al.(2018)Zhu, Thakur, Özaslan, Pfrommer, Kumar and
  Daniilidis}]{zhu_multi_2018}
Zhu AZ, Thakur D, Özaslan T, Pfrommer B, Kumar V and Daniilidis K (2018) The
  multivehicle stereo event camera dataset: An event camera dataset for {3D}
  perception.
\newblock \emph{IEEE Robotics and Automation Letters} 3(3): 2032--2039.

\end{thebibliography}

\end{document}